\documentclass{article}

\usepackage[accepted]{icml2023}
\usepackage{natbib}

\usepackage{float}
    
\usepackage[utf8]{inputenc} 
\usepackage[T1]{fontenc}    
\usepackage{hyperref}       
\usepackage{url}            
\usepackage{booktabs}       
\usepackage{amsfonts}       
\usepackage{nicefrac}       
\usepackage{microtype}      

\usepackage{graphicx}
\usepackage{subfig}
\usepackage{gensymb}
\usepackage{amsthm}
\usepackage{mathrsfs}
\usepackage{amsmath}
\usepackage{amssymb}
\usepackage{mathtools}
\usepackage{wrapfig}
\usepackage{soul}
\usepackage{algorithm}
\usepackage{bbm}
\usepackage{makecell}
\usepackage{listings}

\newcommand{\cdotv}{\boldsymbol{\cdot}}

\usepackage[textwidth=1.2in,textsize=tiny]{todonotes}

\newcommand{\ba}[1]{\begin{align}#1\end{align}}

\usepackage{stackengine}

\makeatletter
\newcommand{\distas}[1]{\mathbin{\overset{#1}{\kern\z@\sim}}}%

\newcommand{\beqs}{\vspace{0mm}\begin{eqnarray}}
\newcommand{\eeqs}{\vspace{0mm}\end{eqnarray}}
\newcommand{\barr}{\begin{array}}
\newcommand{\earr}{\end{array}}

\newcommand{\Pmat}[0]{{{\bf P}}}

\newcommand{\Wmat}[0]{{{\bf W}}}
\newcommand{\Xmat}[0]{{{\bf X}}}

\newcommand{\fv}[0]{{\boldsymbol{f}} }

\newcommand{\wv}{\boldsymbol{w}}
\newcommand{\xv}{\boldsymbol{x}}

\newcommand{\thetav}{\boldsymbol{\theta}}

\newcommand{\muv}[0]{{\boldsymbol{\mu}}}

\newcommand{\given}{\,|\,}

\newcommand{\ours}{{POUF}}

\newcommand{\answerTODO}[1][]{\textcolor{red}{\bf [TODO]}}

\begin{document}

\twocolumn[
\icmltitle{POUF: Prompt-oriented unsupervised fine-tuning for large pre-trained models
}



\icmlsetsymbol{equal}{*}

\begin{icmlauthorlist}
\icmlauthor{Korawat Tanwisuth}{equal,to}
\icmlauthor{Shujian Zhang}{equal,to}
\icmlauthor{Huangjie Zheng}{to}
\icmlauthor{Pengcheng He}{goo}
\icmlauthor{Mingyuan Zhou}{to}
\end{icmlauthorlist}

\icmlaffiliation{to}{The University of Texas at Austin}
\icmlaffiliation{goo}{Microsoft Azure AI}

\icmlcorrespondingauthor{Mingyuan Zhou}{mingyuan.zhou@mccombs.utexas.edu}

\icmlkeywords{Machine Learning, transfer learning, pre-trained models, ICML}

\vskip 0.3in
]
\printAffiliationsAndNotice{\icmlEqualContribution}

\begin{abstract}

Through prompting, large-scale pre-trained models have become more expressive and powerful, gaining significant attention in recent years. Though these big models have zero-shot capabilities, in general, labeled data are still required to adapt them to downstream tasks. To overcome this critical limitation, we propose an unsupervised fine-tuning framework to directly fine-tune the model or prompt on the unlabeled target data. We demonstrate how to apply our method to both language-augmented vision and masked-language models by aligning the discrete distributions extracted from the prompts and target data. To verify our approach's applicability, we conduct extensive experiments on image classification, sentiment analysis, and natural language inference tasks. Across 13 image-related tasks and 15 language-related ones, the proposed approach achieves consistent improvements over the baselines. PyTorch code is available at \url{https://github.com/korawat-tanwisuth/POUF}.

\end{abstract}
\section{Introduction}
The pre-train and fine-tune paradigm has become a standard 
approach to solve many machine learning applications \citep{qiu2020pre, du2022survey}. In this paradigm, a model is first pre-trained on an extensive collection of datasets. While the model may see many examples during pre-training, the target dataset may contain unseen examples with new variations \citep{wenzel2022assaying}. To address the problem of distribution shift between the source and target domains, practitioners fine-tune the pre-trained model on different downstream tasks with task-specific parameters and objective functions. However, this fine-tuning stage generally needs labeled examples, which are expensive to acquire, to adapt the pre-trained model to a specific task.

Without introducing task-specific parameters and objective functions for fine-tuning, recent foundation models, such as CLIP \citep{radford2021learning}, ALIGN \citep{jia2021scaling}, and GPT-3 \citep{brown2020language}, leverage the power of language supervision during pre-training to perform zero-shot predictions through language prompting. For example, to make predictions on a new dataset, users only need to convert class names into textual prompts such as $\text{ ``a photo of a \{CLASS\}.''}$. A prediction is then made by obtaining the prompt yielding the highest similarity score with a given image.  
Despite the capability to
perform zero-shot predictions, the distribution shift problem still persists.

 Recent methods focus on prompt engineering to adapt these models to downstream tasks. Prompt engineering refers to the process of finding the most appropriate prompt to allow a language model to solve the task at hand \citep{liu2021pre}. Many works \citep{wallace2019universal, shin2020autoprompt} focus on discrete prompt search (finding the prompt in the space of the vocabularies of the language model). Since the interpretability of the prompt is often not as important as the performance of the model, various works \citep{li2021prefix, lester2021power, zhou2022learning, zhou2022conditional} propose using continuous prompts that perform prompting in the embedding space of the model. Notably, \citet{lester2021power} and \citet{zhou2022learning} introduce continuous parameters in the word-embedding space and optimize them by maximizing the likelihood of the model on the labeled target data. While these approaches successfully adapt the pre-trained models, they still require a few annotated samples.

\begin{figure*}
[t!] 

\centering
  \subfloat[]{%
        \includegraphics[width=0.21\textwidth]{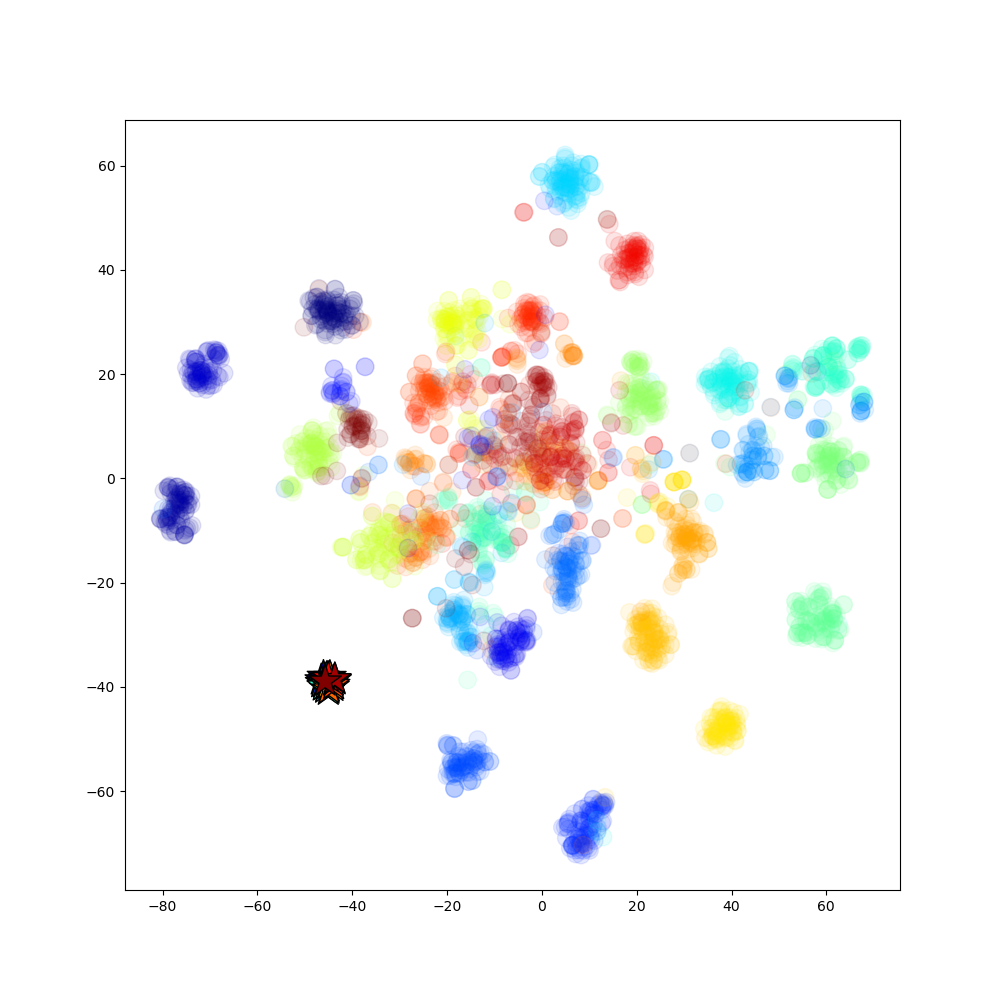}%
        \label{fig:param}%
        }%
    \subfloat[]{%
        \includegraphics[width=0.21\textwidth]{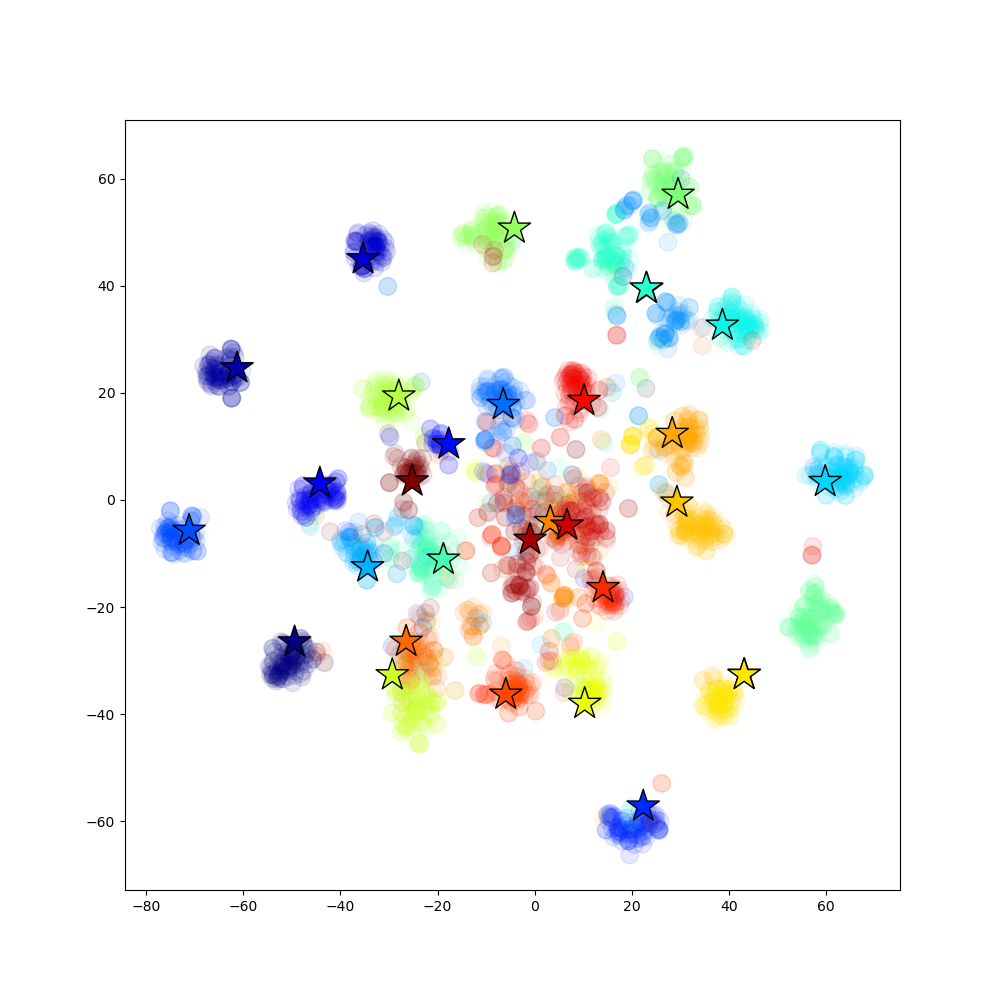}%
        \label{fig:convergence}%
        }%
\subfloat[]{%
        \includegraphics[width=0.21\textwidth]{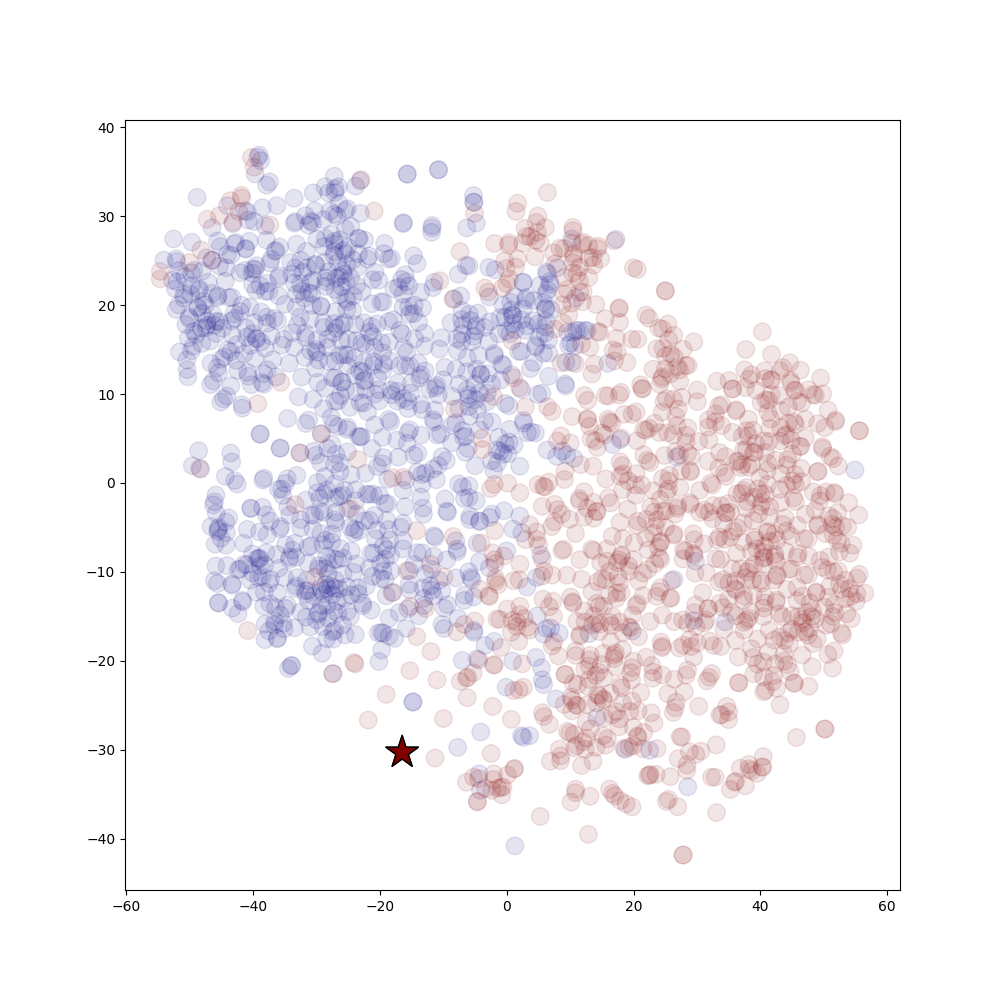}%
        \label{fig:param}%
        }%
    \subfloat[]{%
        \includegraphics[width=0.21\textwidth]{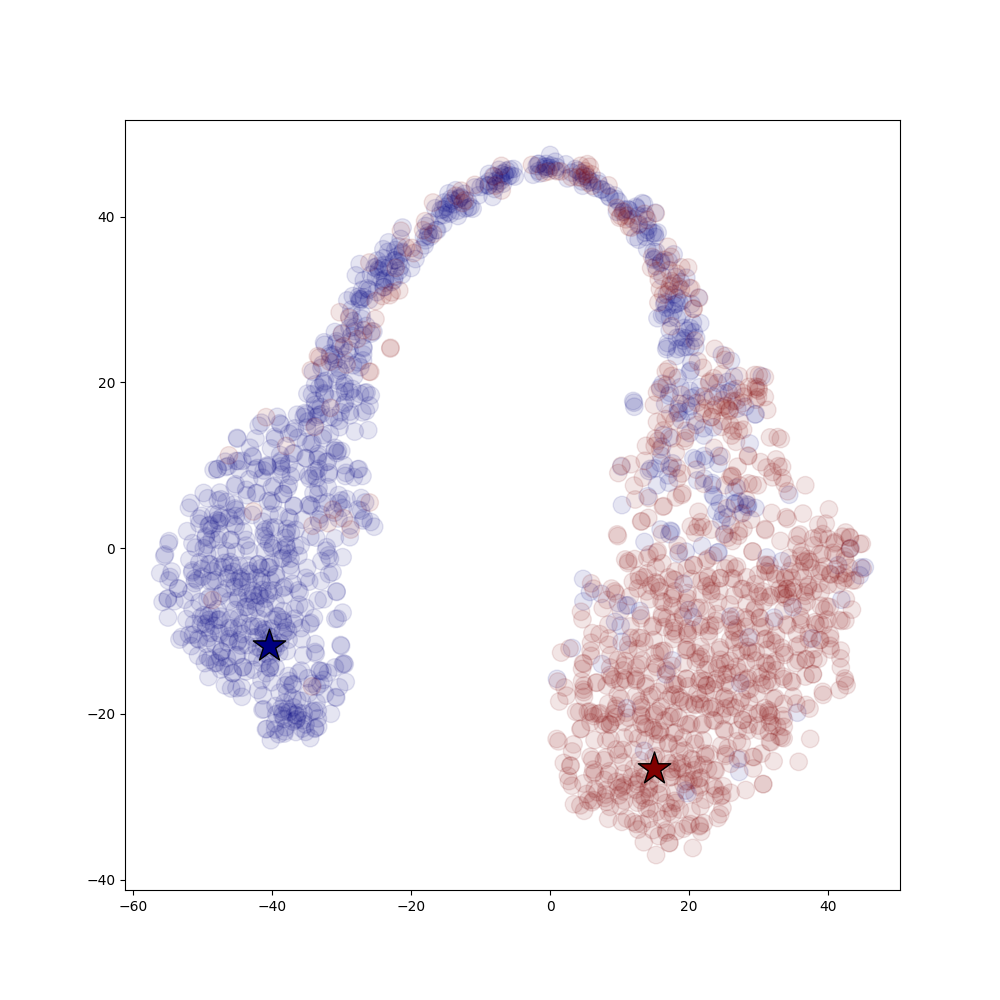}%
        \label{fig:convergence}%
        }%
     \caption{\small t-SNE
visualizations of the embeddings of both the CLIP and RoBERTa models before and after applying \ours, where ($\star$) indicates textual prototypes while ($\cdotv$) represents target features. Different colors correspond to different classes. \textbf{(a)} shows the features and prototypes of the pre-trained zero-shot CLIP model on the Office-31 (Amazon) dataset, while \textbf{(b)} shows those obtained after fine-tuning CLIP with \ours. \textbf{(c)} shows the features and prototypes of the pre-trained zero-shot RoBERTa model on the Subjectivity dataset, while \textbf{(d)} shows those obtained after fine-tuning RoBERTa with \ours. The prototypes of the zero-shot models are not well aligned with the target features as they are clustered around a single point far away from the target features for both CLIP and RoBERTa models. After adapting with \ours, the prototypes and features of their respective classes are well aligned.
} \label{fig:tsne} 
\end{figure*}

To overcome this limitation, we propose prompt-oriented unsupervised fine-tuning (\ours), a simple yet effective framework for fine-tuning pre-trained prompt-based large models with zero-shot capabilities, directly on the unlabeled target data. We formulate unsupervised fine-tuning  as a process of minimizing the statistical distance between the empirical distribution of the textual prototypes and that of the target features. Our framework relies on language prompts to construct class prototypes or target features depending on the model type. For language-augmented vision models, we align the representations of class-specific language prompts, which are class prototypes, with the target image features in the latent space. For masked-language models, we extract the masked-token representations from language prompts and align them with the textual prototypes from the decoder head of the language model. By aligning
these distributions, the pre-trained model can better capture the variations in the target data. To this end, we utilize transport-based alignment and mutual-information maximization objective functions to align the latent representations. Our proposed method is compatible with both full model tuning and prompt tuning.

Our contributions include the following: \textbf{1)} We propose a prompt-oriented framework for fine-tuning pre-trained models with zero-shot capabilities, directly on the unlabeled target data. \textbf{2)} We illustrate how to formulate \ours\ under both language-augmented vision models and masked-language models and demonstrate its effectiveness in practical tasks, such as image classification, sentiment analysis, and natural language inference. \textbf{3)} We perform extensive ablation studies to justify the design decisions of our approach.

\begin{figure*}[hpt!]
    \centering
        \includegraphics[width=.75\textwidth]{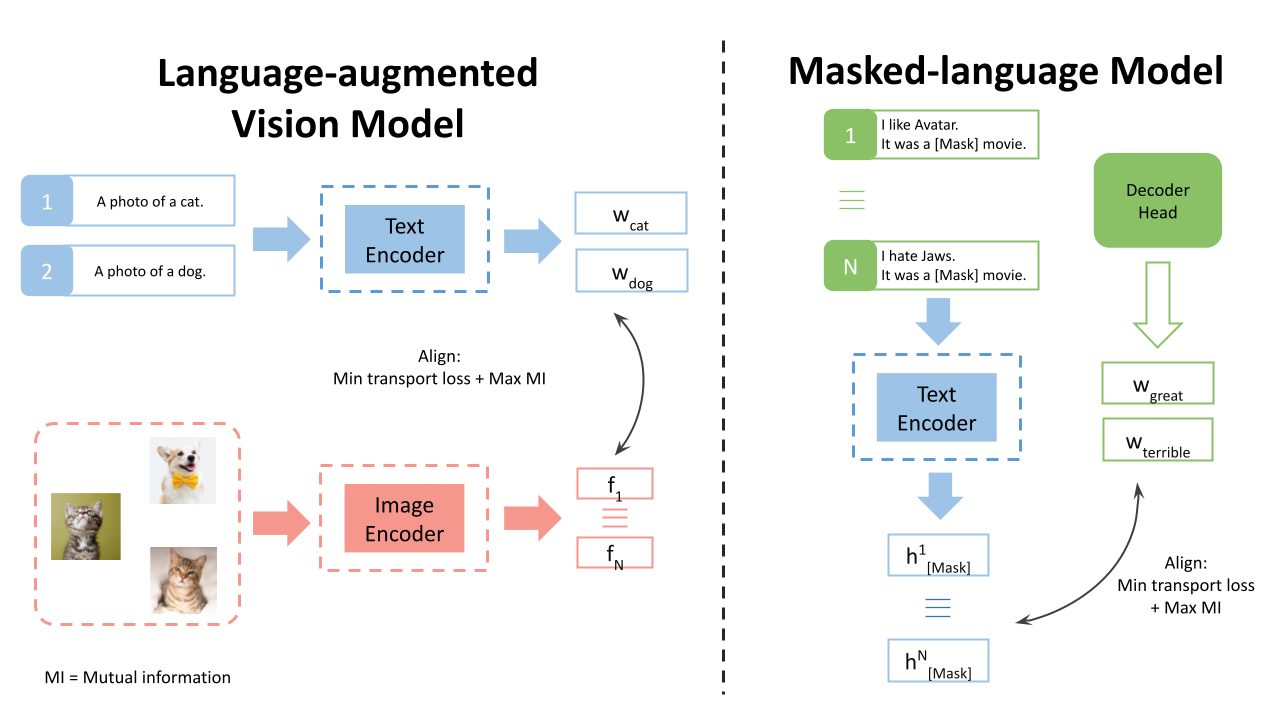}%
        \vspace{-4mm}
        \caption{\small A schematic diagram of \ours\ for language-augmented vision models and masked-language models. Our method addresses the distribution shift problem by aligning the class prototypes with the target features. To this end, \ours\ minimizes the transport cost between the prototypes and target features while maximizing the mutual information between them. For language-augmented vision models, as shown on the left, the textual prototypes are the representations of the prompts. For masked language models, as shown on the right, the textual prototypes are extracted from the decoder head of the underlying language model.
       } 
        \label{fig:diagram}%
\end{figure*}

\section{Prompt-oriented unsupervised fine-tuning}
Below we provide a simple recipe for fine-tuning zero-shot models on unlabeled target data for classification tasks. Our aim is to reduce the distribution shift between the data used to pre-train zero-shot models and the target data. Motivated by the observation that the class prototypes represent source domain information (data for pre-training), we propose to align them with the target representations in the latent space. We demonstrate how to formulate \ours\ for both language-augmented vision and masked-language models. Figure \ref{fig:tsne} provides motivating examples of the latent features learned using our method on different models. It is evident that before applying our method, the textual prototypes are not well aligned with target features, whereas after applying {\ours}, the prototypes and target features become well aligned. A schematic diagram of \ours\ is shown in Figure~\ref{fig:diagram}.

\subsection{\ours\ for language-augmented vision models}
Given an input image $\xv$ from a target dataset $\mathcal{X}$, the CLIP model encodes it through an image encoder to obtain a latent representation as $\fv=F(\xv)$. Each class name is included in a prompt template as an input, $x_{prompt}^k$, to the text encoder, $G(\cdot)$, to obtain textual representation $G(x_{prompt}^k) = \wv_k$ for class $k$. \citet{radford2021learning} 
suggest using the template as $\text{ ``a photo of a \{CLASS\}.''}$ in CLIP as it leads to better performance than using the class names alone. The prediction probabilities over classes are then given as
\begin{align}
P(y=k \mid \xv)=\frac{\exp \left(\cos \left(\fv, \wv_k\right) / T\right)}{\sum_{k'=1}^K \exp \left(\cos \left(\fv, \wv_{k'}\right) / T\right)},
\end{align}
where $T$ is a temperature parameter.

\subsubsection{Textual-prototype construction}
While the model has seen a large collection of data during pre-training, it may not capture the variation in the target data, leading to a sub-optimal performance at the test time without any adaptation. For example, the model may see an image of a polar bear on a white background during pre-training but fail to correctly identify another photo of a polar bear on an unseen background. If we regard the pre-training data as the source domain data, we can apply domain adaptation methods to bridge the domain gap. 

Existing methods in domain adaptation align the distribution of source with that of the target data \citep{tzeng2014deep,ganin2015dann,long2015dan,long2017jan, tzeng2017adversarial, long2018cdan,  zhang2021alignment}. However, this approach presents three challenges. \textbf{1)} Computational inefficiency: The amount of pre-training data is enormous. As an illustration, the CLIP model is trained with 400 million image and text pairs \citep{radford2021learning}. \textbf{2)} Privacy issue: Access to the pre-training data can be restricted. \textbf{3)} Class mismatch: Pre-training data is likely to contain classes that are not present in the target data \citep{cao2018partial}. 

To overcome these challenges, we propose constructing a prototype for each class in the latent space to represent the source data (pre-training data). By using these prototypes to represent source images, the computation is now feasible as there are as many textual prototypes as  classes, whose number is considerably smaller than the number of pre-training data. This strategy also solves the access issue of the pre-training data because we only need to come up with the class names to obtain these textual prototypes. Finally, the target users have control over which classes they want to classify, meaning that there will not be class mismatching. 

Prior works use average latent feature \citep{snell2017prototypical, pan2019transferrable, yue2021prototypical} or learnable weight vectors \citep{saito2019semi, tanwisuth2021prototype, tanwisuth2023prototype} to construct prototypes. The former is computationally expensive as we need to perform multiple forward passings to construct reliable prototypes, while the latter is not applicable since we do not have labeled data. Unlike these approaches, we utilize the textual representations $G(x_{prompt}^k)=\wv_k$, which represent each class in the latent space, as our prototypes. The textual representations will be close to the images of their respective categories. For example, the textual representation of the prompt ``A photo of a cat.'' will be close to cat images. Thus, they represent the prototypes of the classes. To align the distribution of textual prototypes and target data, we first discuss how to construct these distributions and then present different options for distribution alignment.

\subsubsection{Aligning prototypes with target data }
\label{sec:alignment}

 Denoting $\delta$ as the Dirac delta function, we express the distributions over textual representations and target data as 
\ba{\textstyle P=\sum_{k=1}^K u_k \delta_{\wv_k},~~Q=\sum_{i=1}^N v_i \delta_{\fv_i},\label{eq:PQ}} where both $\boldsymbol{u}\in\mathbb{R}_+^K$  and $\boldsymbol{v}\in\mathbb{R}_+^N$
have non-negative elements that sum to one. Unless specified otherwise, we assume $u_k=1/K$ for $k=1,\ldots,K$ and $v_i=1/N$ for $i=1,\ldots,N$ in what follows. To align distributions $P$ and $Q$, we present two alternative methods and discuss the benefits and drawbacks of each.

\textbf{Comparing two distributions.}
Our objective is to align the distribution of textual prototypes and that of the target data. To achieve this, we need to quantify the difference between two discrete distributions. 

One principled approach is to consider Optimal Transport (OT) \citep{villani2008optimal}.
Given two discrete distributions $P$ and $Q$  shown in \eqref{eq:PQ}, 
the OT between them is defined~as
\ba{\operatorname{OT}(P, Q):=\min _{\mathbf{T} \in \Pi(\boldsymbol{u}, \boldsymbol{v})} \operatorname{Tr}\left(\mathbf{T}^{\top} \boldsymbol{C}\right),
\label{eq:OT}
}
where $\mathbf{T} \in \mathbb{R}_{+}^{K \times N}$ is a doubly stochastic transport matrix such that $\Pi(\boldsymbol{u}, \boldsymbol{v})=\left\{\mathbf{T} \mid \mathbf{T} \mathbf{1}_{N}=\boldsymbol{u}, \mathbf{T}^{\top} \mathbf{1}_{K}=\boldsymbol{v}\right\}$, $T_{i j}$ is the transport probability between $x_i$ and $y_j$, $\boldsymbol{C} \in \mathbb{R}_{+}^{K \times N}$ is the transport cost matrix with $C_{ki}=c\left(\wv_k, \fv_i\right)$, and $\operatorname{Tr}(\cdot)$ denotes the matrix trace. 
However, OT can be sensitive to outliers due to the two marginal constraints \citep{chizat2018unbalanced}. Moreover, solving \eqref{eq:OT} 
without any relaxation is a linear programming problem, which has a complexity of $\mathcal{O}(M^3\log M)$, where $M$ denotes the size of the mini-batch of the target examples \citep{merigot2016discrete}. This complexity makes it unfitting for deep learning applications. {\citet{cuturi2013sinkhorn} introduces Sinkhorn divergence, an entropy-regularized OT, to speed up the optimization. While it significantly lowers the computation, it still relies on a two-stage optimization strategy to apply to deep learning models: solving for the transport plan and then updating the network.}

To overcome this computational issue while being deep-learning friendly, \citet{zheng2021exploiting} have introduced the Conditional Transport (CT) framework. Instead of solving for the OT plan, CT defines a bi-directional transport plan, whose two directions correspond to softmax probabilities normalized across the classes and across the data, respectively. This choice makes it easy to integrate it into the deep-learning framework while also effectively aligning the two distributions \citep{tanwisuth2021prototype,wang2022representing}. The CT between two discrete distributions can be written as
\ba{ 
 \mathrm{CT}(P, Q):= \mathcal{L}_{t \rightarrow \wv} + \mathcal{L}_{\wv \rightarrow t},  
}
where $\mathcal{L}_{t \rightarrow \wv}$ is the transport cost from target features to textual prototypes and $\mathcal{L}_{\wv \rightarrow t}$ is the transport cost in the opposite direction. The transport cost from target features to prototypes can be expressed as
\ba{ \mathcal{L}_{t \rightarrow \wv} &= \mathbb{E}_{\xv_i\sim \mathcal{X}} \mathbb{E}_{\wv_k \sim \pi_{\thetav}(\wv_{k}\given  \fv_{i}) }\left[  c(\wv_{k},\fv_{i})\right] \nonumber \\ &~
{ =\mathbb{E}_{\xv_i\sim \mathcal{X}}\left[\sum_{k=1}^K  c(\wv_{k},\fv_{i})
\pi_{\thetav}(\wv_{k}\given  \fv_{i})
\right]}  ,} where $c(\wv_{k},\fv_{i})=1-\cos(\wv_k, \fv_i)$ is the point-wise transport cost, $\pi_{\thetav}(\wv_{k}\given  \fv_{i})=\frac{ p(\wv_k)\exp(\wv_k^T\fv_i)}{\sum_{k'=1}^Kp(\wv_{k'})\exp(\wv_{k'}^T\fv_i  )}$, and $p(\wv_k)$ is the prior probability of prototype $k$.

If we regard the prototypes as the modes of the target distribution, this transport direction has a mode seeking effect, meaning that target features will be close to some prototypes. This direction alone, however, could lead to a degenerate solution where target features are close to only a few prototypes (mode collapse). To counteract this potential issue, CT introduces the transport cost in the opposite direction. 
\ba{ &\resizebox{0.98\hsize}{!}{$\mathcal{L}_{\wv \rightarrow t} = \mathbb{E}_{\{\xv_i\}_{i=1}^N\sim \mathcal{X}}\mathbb{E}_{\wv_k\sim p(\wv_k)} \mathbb{E}_{\fv_i \sim \pi_{\thetav}(\fv_i\given\wv_k  )} \left[ c(\wv_{k},\fv_{i})\right]$} \nonumber \\
	& 
 { =\mathbb{E}_{\{\xv_i\}_{i=1}^N\sim \mathcal{X}}\left[\sum_{k=1}^K p(\wv_k)\sum_{i=1}^N c(\wv_{k},\fv_{i})
 \pi_{\thetav}(\fv_i\given\wv_k )
 \right]}, 
	}
where $\pi_{\thetav}(\fv_i\given\wv_k )=\frac{\exp(\wv_k^T\fv_i)}{\sum_{i'=1}^N\exp(\wv_{k}^T\fv_{i'})}$. In contrast to $\mathcal{L}_{t \rightarrow \wv}$ which has a mode-seeking effect, this transport direction has a mode-covering effect, meaning that each prototype will get target features assigned close to it. Thus, we can avoid the possible mode collapse. We note the definition of modes differs from that in \citet{zheng2021exploiting}, in which one can find a more detailed discussion regarding the mode-covering and mode-seeking behaviors of CT.

The prior term, $p(\wv_k)$, can be set to $\frac{1}{K}$ (a uniform distribution over the classes). However, this approach is not ideal if the class proportions are not balanced. To address the potential class-imbalanced problem, we can learn the prior term from the target data. We discuss an estimation strategy in Appendix \ref{appendix:proportion_esimation}.

Compared to OT, CT has a lower complexity of $\mathcal{O}(d_fMK)$, where $d_f$ is the dimension of the target feature, $M$ denotes the mini-batch size, and $K$ refers to the number of categories. {The optimization of CT is also end-to-end as both the cost and transport plan are parameterized by deep neural networks.}

While \ours\ is compatible with both OT and CT, we empirically find that CT leads to better performance. We show this result in the ablation study in Section \ref{sec:ablation}. The CT-based transport cost is thus expressed as:
\ba{ \mathcal{L}_{transport}\left(F, G; \mathcal{X}\right) = CT(P,Q).}

\subsubsection{Decision-boundary refinement}

Because of the distribution shift problem, some samples may lie close to decision boundaries, meaning that they are far away from the prototypes. Motivated by the cluster assumption \citep{grandvalet2005semi}, which states that decision boundaries should not cross high-density data regions, we propose incorporating another module to refine the predictions of the model on the target data. If the violation of the cluster assumption is minimized, the input examples will be close to the textual prototypes in the latent space.

\textbf{Mutual-information maximization.}
Many existing works in domain adaptation minimize the conditional entropy of the target predictions \citep{vu2019advent,  saito2019semi, saito2020universal, wang2020tent}. However, this has a major shortcoming since this objective alone could lead to a degenerate solution (all samples are clustered around one prototype) \citep{morerio2017minimal, wu2020entropy}. To overcome this issue, we utilize the information maximization objective \citep{krause2010discriminative, shi2012information, liang2020we}. This objective adds a regularization term to conditional entropy minimization. This regularization encourages the model to maximize the marginal entropy of the label distribution \cite{zhang2021capturing}, making the predictions globally diverse while individually certain. The mutual information objective has the following form:
\ba{&\mathcal{L}_{ mi}\left(F, G; \mathcal{X}\right) =-[H\left(\mathcal{Y}\right)-H\left(\mathcal{Y} \mid \mathcal{X}\right)]  \nonumber \\ &~~~~~~=-[h\left(\mathbb{E}_{\xv \in \mathcal{X}} p(y| \xv)\right)-\mathbb{E}_{\xv \in \mathcal{X}} h\left(p(y| \xv)\right)], \label{eq:mi_loss}}
where $H\left(\mathcal{Y}\right)$ and $H\left(\mathcal{Y} \mid \mathcal{X}\right)$ denote the marginal entropy and conditional entropy of the text categories $\mathcal{Y}$, respectively, and $h(p) = -\sum_i p_i\log p_i$.

Putting it all together, we write the final loss function as
\ba{\resizebox{0.92\hsize}{!}{$\mathcal{L}_{LVM}\left(F, G; \mathcal{X}\right)=\mathcal{L}_{transport}\left(F, G; \mathcal{X}\right)+\lambda \mathcal{L}_{ mi}\left(F, G; \mathcal{X}\right), $}\!\! \label{eq:final_loss}}
where $\lambda$ is a hyper-parameter controlling the weight of the mutual-information objective. We then update both the image encoder, $F$, and the text encoder, $G$, through gradient backpropagation of this loss function. Alternatively, we can introduce soft-prompt parameters in the word embedding space and update only these parameters while keeping the model parameters fixed. We discuss this alternative design consideration in more detail in Section \ref{sec:design_considerations}.

\subsection{\ours\ for masked-language models}
In a conventional setting, to fine-tune masked-language models such as BERT \citep{devlin2018bert} and RoBERTa \citep{liu2019roberta} on a task, one first converts input $x_1$ to a sequence of token~$\tilde{x}$. The language model, M, then maps $\tilde{x}$ to a sequence of hidden vectors $\{h_k \in \mathbb{R}^d\}$. One then introduces a task-specific head $\Wmat_0\in \mathbb{R}^{|\mathcal Y|\times d}$, which is randomly initialized, to classify the hidden representations through $\text{softmax}(\Wmat_0 \mathbf{h}_{[\text{CLS}]})$ into a class in $\mathcal{Y}$. Since $\Wmat_0$ does not capture any semantic meaning,   it is trained together with the parameter of the pre-trained model by minimizing the negative log probability over the labels $y \in \mathcal{Y}$.
The drawbacks of this approach are that one needs to not only introduce additional task-specific parameters but also utilize labeled examples to fine-tune the model.

An alternative way to fine-tune the language model on this task is to formulate it as a masked-language modeling problem, mimicking the pre-training process \citep{schick2020exploiting, gao2020making}. During this process, one first maps the labels to the words in the vocabulary. As an example, for a sentiment classification task, a positive label ($+$) can be mapped to the word ``great'' whereas a negative label ($-$) can be mapped to the word ``terrible''. The language model is then tasked to fill the mask token with the label words (``great'' or ``terrible''). Given an input $x_1$, one can then construct a prompt from the text input as:
$$x_{prompt} = [\text{CLS}]\ x_1\ \text{It was [MASK]. [SEP]}.$$ 

The prompt can be manually or automatically generated, so long as the input contains a mask token. Having a mask token in the input allows one to treat the problem as a masked language modeling task, $i.e.$,  directly modeling the probability of each label as that of the corresponding word in  the language model's vocabulary as
\begin{align}
P\left(y \mid x_{\mathrm{in}}\right) &=p\left([\text { MASK }]=\mathcal{M}(y) \mid x_{\mathrm{prompt}}\right) \\
&=\frac{\exp \left(\mathbf{w}_{\mathcal{M}(y)} \cdot \mathbf{h}_{[\text {MASK }]}\right)}{\sum_{y^{\prime} \in \mathcal{Y}} \exp \left(\mathbf{w}_{\mathcal{M}\left(y^{\prime}\right)} \cdot \mathbf{h}_{[\text {MASK }]}\right)},
\end{align}
where $\mathbf{w}_{v}$ corresponds to the weight vector of the vocabulary $v \in \mathcal{V}$ and $\mathbf{h}_{[\text{MASK}]}$ denotes the hidden representation of the mask token, and $\mathcal{M}:\mathcal{Y}\rightarrow \mathcal{V}$ represents the mapping from the set of labels to the words in the vocabulary. 

\subsubsection{Textual-prototype construction} To apply \ours\ to masked-language models, we first construct the textual prototypes, which contain information about the classes. Unlike language-augmented vision models' prompts which encode class information, Masked-language models' prompts contain information about the input examples. Thus, we cannot use them as our class prototypes. To construct textual prototypes, we propose using the weight of the last layer of the decoder head, which is tasked to predict the mask tokens during pre-training. These weights capture the meanings of the individual words in the vocabulary of the language model. For example, $\wv_{\text{great}}$ will be close to the masked-token representations of the sentences which contain the word ``great''. We then use the alignment module to align \ba{
\textstyle
P=\sum_{y\in \mathcal{Y}}\frac{1}{|\mathcal{Y}|}\delta_{\wv_{\mathcal{M}(y)}} ~\text{ and }~Q=\sum_{i=1}^N\frac{1}{N}\mathbf{h}_{\text{[MASK]}}^i.} The information maximization objective is directly computed via the predictive distribution of the language model. 

Putting it all together, we write the final loss function as
\ba{\resizebox{0.9\hsize}{!}{$\mathcal{L}_{MLM}\left(M; \mathcal{X}\right) =\mathcal{L}_{transport}\left(M ; \mathcal{X}\right)+\lambda \mathcal{L}_{ mi}\left(M; \mathcal{X}\right),$}\!\! \label{eq:final_loss}}
where $\lambda$ is a hyper-parameter controlling the weight of the mutual-information objective.

\subsection{Design considerations}
\label{sec:design_considerations}
As briefly discussed in the previous section, we could perform full model tuning or prompt tuning. \textbf{1)} {\it Full-model tuning} refers to updating the parameters of both the text and image encoders for language-augmented vision models and the text encoder for masked-language models. This design generally leads to better performance but requires more resources for tuning. 
\textbf{2)} {\it Prompt tuning \citep{lester2021power}} refers to introducing additional tunable parameters in the input embedding space. Although using fewer resources for fine-tuning, it may not necessarily lead to optimal performance. 

We now give a more formal description of prompt tuning. Given a sequence of $n$ tokens, $\{x_1,x_2, \ldots, x_n\}$, a transformer-based language model maps this sequence to a matrix $\Xmat_h \in \mathbb{R}^{n\times h}$ where $h$ is the hidden dimension. Instead of tuning the model parameter, prior works \citep{lester2021power, zhou2022learning} introduce soft-prompt parameter $\Pmat_h \in \mathbb{R}^{p \times h}$ where $p$ is the prompt length. The embedded prompt and input are then concatenated to form a matrix $[P_h; X_h] \in \mathbb{R}^{(p+n) \times h}$. This matrix then goes through the encoder, but only the parameters of the soft-prompts $\Pmat_h$ are updated.

\section{Related work}

We summarize how POUF differs from related work in Table~\ref{tab:subof} and provide more details below.

\begin{table}[t] 
\centering
\caption{\small Data and model requirements for different transfer learning settings. POUF requires neither source data nor target labels.}
\vspace{-3mm}
 \resizebox{\columnwidth}{!}{\begin{tabular}{ccccc}
\toprule Setup & Source data  & Target data & Parameter update \\
\midrule
 \text {Fine-tuning } & \text { N/A} & \text { Labeled}  & Full model \\
  \text {Domain adaptation } & \text { Labeled} & \text { Unlabeled}  & Full model \\
   \text {Prompt tuning } & \text { N/A} & \text { Labeled }  & Prompt \\
   \midrule
\makecell {POUF} & \text { N/A} & \text { Unlabeled }  & Prompt/full model \\
\bottomrule
\end{tabular}} \vspace{-3mm}
\label{tab:subof}
\end{table}

\textbf{Learning under distribution shift.}
Traditional domain adaptation methods jointly optimize on labeled source data and unlabeled target data \citep{tzeng2014deep,ganin2015dann,long2015dan,long2017jan, tzeng2017adversarial, long2018cdan,  zhang2021alignment, zhang2021learning}. The requirement for labeled source data limits their use in many applications. \ours, however, can directly adapt to the target domain data, greatly increasing its general applicability. This also means that the performance of domain adaptation methods heavily depends on the source dataset, whereas \ours\ is source domain agnostic. Another related line of work is ``source-free'' domain adaptation \citep{liang2020we, li2020model,kundu2020class,kundu2020universal, kurmi2021domain}. In this paradigm, the pre-trained model is adapted in two steps. First, it is fine-tuned on labeled source data. Then, it is adapted to the target data without accessing the source data. In this sense, they are not completely source-free. Different from this line of work, \ours\ leverages textual prototypes in models with zero-shot capabilities to adapt to the target data directly, bypassing the fine-tuning step on the labeled source data.

\textbf{Prompt-based learning.}
For a full review of this topic, we refer the readers to the survey by \citet{liu2021pre}. Several works that focus on prompt tuning can be broadly classified into two categories: hard and soft prompt tuning. Soft prompt tuning \citep{li2021prefix, lester2021power, zhou2022learning, zhou2022conditional} introduces continuous parameters in the word embedding space and tunes these parameters instead of the full model, whereas hard prompt tuning \citep{wallace2019universal, shin2020autoprompt, zhang2022allsh} searches for discrete tokens in the vocabulary of the language model to optimize the performance of the model. These works focus on prompt design and engineering and rely on labeled data. Unlike these methods, ours is unsupervised, making it more generally applicable. However, as we show in the experiments, our work is complementary to these approaches since we can apply these methods to obtain better prototypes and then further fine-tune using our framework. Recently, \citet{huang2022unsupervised} propose a method for unsupervised prompt tuning for language-augmented vision models. Different from this work, our work provides a general framework for both language-augmented vision and masked-language models. 

\textbf{Prototype-based learning.} The idea of using prototypes has been studied in few-shot classification \citep{snell2017prototypical,guo2022learning}, self-supervised learning \citep{asano2019self, caron2020unsupervised, li2020prototypical}, and domain adaptation \citep{pan2019transferrable,kang2019contrastive,liang2020we, tanwisuth2021prototype}.
In these works, the prototypes are defined as the average latent features or learnable weight vectors. Different from them, we propose using textual representations to represent our prototypes. This is the key idea of our method as it allows us to adapt directly to the target data.

\section{Experiments}
\label{sec:experiments}
We evaluate POUF on both language-augmented vision and masked-language models and compare it to a rich set of baselines in various vision and language modeling tasks.
\subsection{\ours\ for language-augmented vision models}

\textbf{Datasets.} 
\textbf{1)} {\it Office-31 \citep{saenko2010adapting}} contains 4,652 images with 31 classes from three domains: Amazon (A), Webcam (W), and DSLR (D). \textbf{2)} {\it Office-Home  \citep{venkateswara2017deep}} has 15,500 images with 65 classes from four domains: Artistic images (Ar), Clip art (Cl), Product images (Pr), and Real-world (Rw), making it more challenging than Office-31. \textbf{3)} {\it DomainNet \citep{peng2019moment}}, a large-scale dataset, consists of 569,010 images with 345 categories from six domains: Clipart, Infograph, Painting, Quickdraw, Real, and Sketch. More details on the datasets can be found in Appendix \ref{appendix:datasets}.

\textbf{Baselines.}
\textbf{1)} {\it Clip (zero-shot) \citep{radford2021learning}} refers to using the CLIP model to perform zero-shot predictions on the target dataset. \textbf{2)} {\it Tent \citep{wang2020tent}}  refers to adapting the model with entropy minimization before predicting on the target dataset. We note that we have slightly modified Tent since the vision transformer (ViT) architecture does not have modulation parameters (batch normalization). Thus, we introduce soft-prompt parameters and update them instead.  \textbf{3)} {\it Unsupervised prompt learning (UPL) \citep{huang2022unsupervised}} refers to using the top-K confident predictions for each class as pseudo labels. After obtaining the pseudo labels, we train the soft-prompt parameters with the cross-entropy loss. For a fair comparison, we do not use model ensemble as done in \citet{huang2022unsupervised}. To help understand the gap between unsupervised methods and supervised ones, we also provide the results of a representative few-shot learning method, \textbf{4)} {\it Context Optimization \citep{zhou2022learning}} (CoOp). We note that this baseline should not be directly compared to the unsupervised ones, but rather serves as a reference to understand how much labeled data can help boost performance.

\textbf{Implementation details.}
We build our method using the open-source CLIP codebase \citep{radford2021learning} and TLlib transfer learning library \citep{jiang2022transferability}. For all experiments, we adopt the ViTB-16 for the image encoder and the default transformer from the CLIP paper for the text encoder. All the unlabeled target samples are used for fine-tuning. The learning rate schedule is set to $\eta_{\text{iter}}=\eta_0(1+\gamma\text{iter})^{-\alpha}$, where $\eta_0$ is the initial learning rate. We adopt the following default hyper-parameters: $\gamma=2\mathrm{e}{-4}$, and $\alpha=0.75$. We set $\eta_0=5\mathrm{e}{-7}$ for all experiments except for prompt tuning on Office-31 where $\eta_0=1\mathrm{e}{-3}$. We use a mini-batch SGD with a momentum of $0.9$ and a batch size of $96$ for Office-31 and Office-Home and $16$ for DomainNet. The weight of the mutual-information objective, $\lambda$, is set to $0.3$ for all experiments. 

\textbf{Main results for language-augmented vision models.}
We conduct systematic experiments on 13 image tasks with varying data sizes. In each experiment, we fine-tune the CLIP model with unlabeled target images. Table \ref{tab:main_image} exhibits the results of \ours\ and the baselines. Tent and UPL both achieve consistent gains over the zero-shot CLIP model on both the Office-31 and Office-Home datasets. However, on the more challenging DomainNet, both baselines suffer from negative transfers. By contrast, \ours\ improve the CLIP model by $11.8$\%, $3.7$\%, and $3.6$\% on the Office-31, Office-Home, and DomainNet datasets, respectively. The consistent improvements illustrate that \ours\ is an effective strategy to fine-tune the model with unlabeled target data. Moreover, \ours\ is compatible with both prompt tuning and model tuning as evident in the performance gains. Prompt tuning yields slightly worse performance than model tuning, as we expected. However, its memory footprint is significantly lower. We provide parameter and run-time analyses \cite{fan2020bayesian,zhang2021bayesian} of the two approaches in Appendix \ref{appendix:analysis}. 

POUF with model tuning outperforms the few-shot learning method (CoOp) in 5 out of 13 tasks. This result illustrates that there is still a gap between the supervised and unsupervised methods. However, the benefit of a large amount of unlabeled data can sometimes outweigh the benefit of a small amount of labeled data.

To understand the generalization ability of the model after adapting with POUF, we further conduct an unseen class experiment in Appendix \ref{appendix:unseen}.
POUF yields improved performance for the seen class but slightly weaker generalization for the unseen class. This outcome highlights a minor trade-off between the model's specialization and generalization. However, if target users have access to unseen class names, training POUF with both seen and unseen class text prototypes results in superior performance compared to the zero-shot model for both seen and unseen classes. See Appendix \ref{appendix:unseen} for a detailed analysis. 

\begin{table*}[t!]
\centering
 \caption{\small Accuracy $(\%)$ on three different datasets for methods based on CLIP. } 
 \vspace{-2mm}
\subfloat[]{
 \resizebox{.8\textwidth}{!}{\begin{tabular}{c|c|ccc|c|cccc|c}
\toprule  Category & Methods & \multicolumn{4}{c|}{Office-31} & \multicolumn{5}{c}{Office-Home}   \\
 \cmidrule(r){3-6} \cmidrule(r){7-11} 
\multicolumn{1}{c}{} & \multicolumn{1}{|c}{} & \multicolumn{1}{|c}{ A} & \multicolumn{1}{c}{ D} & \multicolumn{1}{c}{ W} & \multicolumn{1}{|c|}{Avg} & \multicolumn{1}{c}{ Ar} & \multicolumn{1}{c}{ Cl} & \multicolumn{1}{c}{ Pr} & \multicolumn{1}{c}{ Rw} & \multicolumn{1}{|c}{Avg} \\
\midrule
& CLIP (zero-shot) \citep{radford2021learning} & $79.0$ & $77.5$ & $74.7$ & $77.1$ & $82.7$ & $68.1$ & $89.1$ & $89.8$ & $82.4$ \\

& Tent \citep{wang2020tent} & $81.5(0.1)$	& $80.7(0.7)$ & $82.8(0.2)$	& $81.7$  
 & $83.2 (0.1) $ & $67.8(0.2)$ & $91.9(0.1)$	& $90.4(0.1)$	& $83.3$ \\
Unsupervised & UPL \citep{huang2022unsupervised} & $81.4(0.1)$ & $82.6(0.1)$ & $83.6(0.3)$ & $82.5$ & $83.3(0.1)$ & $67.7(0.1)$ & $91.5(0.1)$ & $90.7(0.1)$	& $83.3$ \\
\cmidrule(r){2-11}
& {\bf\ours}\ (prompt tuning)  & $83.6 (0.5)$& $89.9 (2.5)$& $90.6 (2.3)$& $88.0 $& $83.7 (0.1)$& $71.2 (0.1)$& $91.4 (0.1)$& $90.8 (0.1)$& $84.3 $\\
& {\bf \ours}\ (model tuning)  & $\bf{84.4} (0.4)$& $\bf{91.1} (2.1)$& $\bf{91.3} (2.5)$& $\bf{88.9} $& $\bf{86.2} (0.4)$& $\bf{73.8} (0.3)$& $\bf{92.7} (0.1)$& $\bf{91.7} (0.1)$& $\bf{86.1} $\\
\midrule
Few-shot & CoOp \citep{zhou2022learning} & $83.7(2.1)$ & $98.8(0.6)$ & $98.2(0.8)$ & $93.6$ & $85.6(0.4)$ & $73.2(0.2)$ & $92.7(0.1)$ & $91.6(0.1)$ & $85.8$ \\
\bottomrule
\end{tabular}}
}
 \vspace{-2mm}
\subfloat[]{
 \resizebox{0.7\textwidth}{!}{\begin{tabular}{c|c|cccccc|c}
\toprule  Category & Methods & \multicolumn{7}{c}{DomainNet}  \\
\cmidrule(r){3-9} 
\multicolumn{1}{c}{} & \multicolumn{1}{|c}{} & \multicolumn{1}{|c}{ C} &\multicolumn{1}{c}{ I}& \multicolumn{1}{c}{ P}& \multicolumn{1}{c}{ Q}& 
\multicolumn{1}{c}{ R}& 
\multicolumn{1}{c}{ S}& 
\multicolumn{1}{|c}{Avg}\\
\midrule
& CLIP (zero-shot) \citep{radford2021learning} &  $70.9$ & $48.2$ & $65.9$ & $14.0$ & $83.6$ & $63.6$ & $57.7$ \\
& Tent \citep{wang2020tent} &  $71.4(0.2)$ & $47.8(0.5)$  & $66.2(0.1)$ & $14.2(0.1)$ & $83.9(0.1)$ & $64.1(0.2)$ & $57.9$\\
Unsupervised & UPL \citep{huang2022unsupervised} &  $71.7(0.1)$ & $47.5(0.4)$  & $66.3(0.2)$ & $14.4(0.3)$ & $83.8 (0.1)$ & $64.3(0.1)$ & $58.0$\\
\cmidrule(r){2-9} 
& {\bf \ours}\ (prompt tuning)  & $72.8 (0.1)$& $53.1 (0.3)$& $68.6 (0.1)$& $15.9 (0.1)$& $84.4 (0.0)$& $66.2 (0.1)$& $60.2 $\\
& {\bf \ours}\ (model tuning)  & $\bf{73.8} (0.1)$& $\bf{55.7} (0.3)$& $\bf{68.6} (0.1)$& $\bf{18.8} (0.7)$& $\bf{84.6} (0.0)$& $\bf{66.4 }(0.1)$& $\bf{61.3} $\\
\midrule
Few-shot & CoOp \citep{zhou2022learning} & $75.3(0.2)$ & $55.7(0.2)$ & $71.9(0.2)$ & $20.7(0.2)$ & $83.4(0.1)$ & $67.4(0.1)$ & $62.9$  \\

\bottomrule
\end{tabular}}
} 
\label{tab:main_image}
\end{table*}

\begin{table*}[t!]
\centering
 \caption{\small Results on various datasets for methods based on RoBERTa-large.}
 \vspace{-2mm}
 \resizebox{.97\textwidth}{!}{\begin{tabular}{c|c|cccccccc}
\toprule Category & Methods  & SST-2 (acc)  & SST-5 (acc) & MR (acc) & CR (acc)& MPQA (acc)& Subj (acc) & TREC (acc)  & CoLA (Matt.) \\
\midrule
& Majority & $50.9$ & $23.1$ & $50.0$ & $50.0$ & $50.0$ & $50.0$ & $18.8$ & $0.0$\\
Unsupervised & RoBERTa-large (zero-shot) & $83.6$ & $35.0$ & $80.8$ & $79.5$ & $67.6$ & $51.4$ & $32.0$ & $\bf{2.0}$\\

\cmidrule{2-10}
& {\bf \ours}  & $\bf{89.4} (0.0)$& $\bf{44.9} (1.6)$& $\bf{86.4} (0.9)$& $\bf{86.0} (5.6)$& $\bf{78.1} (7.8)$& $\bf{80.8} (0.1)$& $\bf{36.9} (15.1)$& $1.5 (0.1)$\\
\midrule
& Few-shot fine-tuning & $81.4  (3.8)$& $43.9  (2.0)$& $76.9  (5.9)$& $75.8  (3.2)$& $72.0  (3.8)$& $90.8  (1.8)$& $88.8  (2.1)$& $\bf{33.9}  (14.3)$\\
& “GPT-3” in-context learning & $84.8  (1.3)$& $30.6  (0.9)$& $80.5  (1.7)$& $87.4  (0.8)$& $63.8  (2.1)$& $53.6  (1.0)$& $26.2  (2.4)$& $-1.5  (2.4)$\\
Few-shot & LM-BFF (man) & $92.6  (0.5)$& $50.6  (1.4)$& $86.6  (2.2)$& $90.2  (1.2)$& $87.0  (1.1)$& $92.3  (0.8)$& $87.5  (3.2)$& $18.7  (8.8)$\\
\cmidrule{2-10}
 & {\bf LM-BFF (man) + \ours} & $\bf{93.2} (0.0)$& $\bf{53.6} (0.2)$& $\bf{87.6} (0.2)$& $\bf{91.8} (0.2)$& $\bf{88.0} (0.2)$& $\bf{92.9} (0.1)$& $\bf{90.1} (0.1)$& $20.6 (0.6)$\\
\midrule
&&MNLI (acc) & MNLI-mm (acc) &SNLI (acc) &QNLI (acc) &RTE (acc) &MRPC (F1) &QQP (F1) & STS-B (Pear.)\\
    
\midrule
& Majority & $32.7$ & $33.0$ & $33.8$ & $49.5$ & $52.7$ & $\bf{81.2}$ & $0.0$ & $-$\\
Unsupervised & RoBERTa-large (zero-shot) & $50.8$ & $51.7$ & $49.5$ & $50.8$ & $51.3$ & $61.9$ & $49.7$ & $-3.2$\\

\cmidrule{2-10}
& {\bf \ours} & $\bf{55.0} (2.3)$& $\bf{56.3} (3.4)$& $\bf{64.0} (0.2)$& $\bf{68.2} (4.0)$& $\bf{64.5} (2.5)$& $\bf{81.2} (0.1)$& $\bf{53.8} (0.1)$& $\bf{20.9} (9.3)$ \\
\midrule
& Few-shot fine-tuning& $45.8  (6.4)$& $47.8  (6.8)$& $48.4  (4.8)$& $60.2  (6.5)$& $54.4  (3.9)$& $76.6  (2.5)$& $60.7  (4.3)$& $53.5  (8.5)$\\
& “GPT-3” in-context learning & $52.0  (0.7)$& $53.4  (0.6)$& $47.1  (0.6)$& $53.8  (0.4)$& $60.4  (1.4)$& $45.7  (6.0)$& $36.1  (5.2)$& $14.3  (2.8)$\\
Few-shot & LM-BFF (man) & $70.7  (1.3)$& $72.0  (1.2)$& $79.7  (1.5)$& $69.2  (1.9)$& $68.7  (2.3)$& $77.8  (2.0)$& $\bf{69.8}  (1.8)$& $73.5  (5.1)$\\
\cmidrule{2-10}
& {\bf LM-BFF (man) + \ours} & $\bf{73.8} (1.1)$& $\bf{74.9} (0.1)$& $\bf{80.0} (0.2)$& $\bf{76.3} (0.3)$& $\bf{70.5} (0.6)$& $\bf{81.8} (0.7)$& $69.0 (1.0)$& $\bf{82.1} (0.1)$\\
\bottomrule
\end{tabular}}
\label{tab:main_text}
\end{table*}

\subsection{\ours\ for masked-language models}
\textbf{Datasets.}
We perform experiments on 15 English-language tasks, 8 of which are single-sentence tasks and 7 of which are sentence-pair tasks. These datasets include popular classification tasks such as SST-5, MR, CR, MPQA, Subj, and TREC as well as the GLUE benchmark \citep{wang2018glue}. We provide a table of different tasks in Appendix \ref{appendix:datasets}. Given an input sentence, the goal of the single-sentence task is to predict the label. E.g., given a movie review, we need to predict if it is positive or negative. Similarly, the aim of the sentence-pair task is to predict the relationship between a pair of input sentences. As an illustration, given a premise and a hypothesis, we need to predict whether the hypothesis is an entailment, neutral, or contradiction of the premise. 

\textbf{Baselines.}
We directly take the baselines and results from 
\citet{gao2020making}.
\textbf{1)} {\it Majority} denotes the proportion of the majority class in the data. \textbf{2)} {\it RoBERTa-large (zero-shot)} refers to using prompt-based prediction by RoBERTa-large for zero-shot predictions. \textbf{3)} {\it Few-shot fine-tuning} refers to standard fine-tuning with few-shot examples. \textbf{4)}{\it ``GPT-3'' in-context learning}  refers to prompt-based predictions with augmented context (32 randomly sampled demonstrations). The model is still RoBERTa-large (not GPT-3). \textbf{5)} {\it LM-BFF (man)} refers to the complete method by \citet{gao2020making}, where the prompt is manually selected using the template. 

\textbf{Implementation details.}
We follow the experimental protocols in \citet{gao2020making}. Specifically, for each task, the data is split into $\mathcal{D}_{train}$, $\mathcal{D}_{dev}$, and $\mathcal{D}_{test}$. The authors tune the hyper-parameters on $\mathcal{D}_{dev}$ and report the performance of the model on $\mathcal{D}_{test}$. We take the same hyper-parameters as the original paper: batch size $=8$, learning rate $=1e-5$, training steps $=1,000$, and few-shot examples per class $=16$. The weight of the mutual-information objective, $\lambda$, is set as $0.6$ and the weight for the transport cost is set as one for all experiments except for the unsupervised setting on the MRPC and QQP tasks. In these two tasks, the weights are set to $1\mathrm{e}{-4}$ for the mutual-information objective and to $1\mathrm{e}{-2}$ for the transport cost.
We validate the model's performance every $100$ steps on 
$\mathcal{D}_{dev}$ and take the best validated checkpoint for the final evaluation on $\mathcal{D}_{test}$.

\begin{table*}[t] 
\centering
\caption{\small Average accuracy (\%) of \ours\ on Office-31 (CLIP) and RTE (RoBERTa) under different variants.}
\vspace{-2mm}
 \resizebox{1.0\textwidth}{!}{ \begin{tabular}{c|cccccc}
\toprule
Datasets & OT for $\mathcal{L}_{transport}$ & OT-Sinkhorn for $\mathcal{L}_{transport}$ &
 \ours\ w/o $\mathcal{L}_{transport}$& \ours\  w/o $\mathcal{L}_{mi}$  & $c(\muv_k, \fv_j^t)=\exp(-\muv_k^{T}\fv_j^t)$& \ours\ ({\small default setting}) \\
\midrule
Office-31 & $86.1 \pm 0.2$ & $86.5\pm 0.2$& $83.3 \pm 0.4$ & $84.2 \pm 0.2$ & $87.6\pm 0.5$  & $88.0 \pm 0.9$\\
RTE & $58.0 \pm 5.5$ & $63.3\pm 6.0$& $56.9 \pm 1.5$ & $58.4 \pm 1.3$ & $61.3\pm 6.3$  & $64.5 \pm 2.5$\\
\bottomrule
\end{tabular}} 
\label{tab:ablation}
\end{table*}

\textbf{Main results for masked-language model.}
To validate the effectiveness of \ours\ on language tasks, we fine-tune the RoBERTa-large model with \ours\ on the unlabeled text data. We also incorporate \ours\ into LM-BFF to show the compatibility of our method with few-shot learning frameworks. We report the results on the 15 language tasks in Table \ref{tab:main_text}. Under the unsupervised category, zero-shot RoBERTa model yields consistently higher performance than majority class. This result indicates that the model contains knowledge that can be exploited by prompt-based prediction. After applying {\ours}, the model performance significantly improves on 14 tasks. Under the few-shot category, few-shot fine-tuning does not always lead to a better performance than prompt-based prediction. Similarly, ``GPT-3'' in-context learning sometimes hurts the prompt-based prediction performance. We speculate that the size of the language model may have an impact on the performance. Since \citet{gao2020making} use RoBERTa-large instead of the original GPT-3 model, the smaller size of RoBERTa may have a significant impact on the performance of in-context learning. While \ours\ does not rely on labeled data, it outperforms both few-shot fine-tuning and ``GPT-3'' in-context learning on 10 and 14 tasks, respectively. 
For the few-shot setting, we incorporate POUF into LM-BFF by using the few-shot examples without labels to compute POUF's loss and add it to LM-BFF's objective. The results show that
\ours\ also
consistently boosts the performance of LM-BFF across 14 out of 15 tasks, illustrating the compatibility of our method with few-shot learning frameworks.

\subsection{Analysis of results}
\label{sec:ablation}

\textbf{Ablation studies.}
To verify the effect of each component of our method, we conduct ablation studies on both the Office-31 (image) and RTE (language) datasets and report the results in Table~\ref{tab:ablation}. \textbf{1)} {\it Distribution-matching options.} We provide two alternatives for CT: OT and OT-Sinkhorn. OT solves the exact linear program to obtain the optimal couplings, while OT-Sinkhorn solves a relaxation of the OT problem with the Sinkhorn algorithm \cite{cuturi2013sinkhorn}. In each variant, we replace CT in the $\mathcal{L}_{transport}$ while keeping $\mathcal{L}_{mi}$. In both image and language tasks, \ours\ with CT outperforms \ours\ with OT and OT-Sinkhorn. \textbf{2)} {\it Significance of each loss.} We remove each loss from the framework to understand the effect of each part. Without the transport cost, the performance drops by $4$\% and $7$\% on the image and text tasks, respectively. Similarly, after removing the mutual-information objective, the accuracy decreases by $4$\% and $6$\% on the image and text tasks, respectively. These results illustrate the significance of each loss and the synergistic effect of the two losses. \textbf{3)} {\it Cost function.} In the transport framework, we utilize the cosine distance as the cost function. Alternatively, we could use other cost functions. We experiment with another cost function inspired by the radial basis kernel. The result shows that the cosine distance function leads to better performance in both modalities, justifying our design decisions.    

\begin{figure}[t!]
    \centering
        \includegraphics[width=0.5\textwidth]{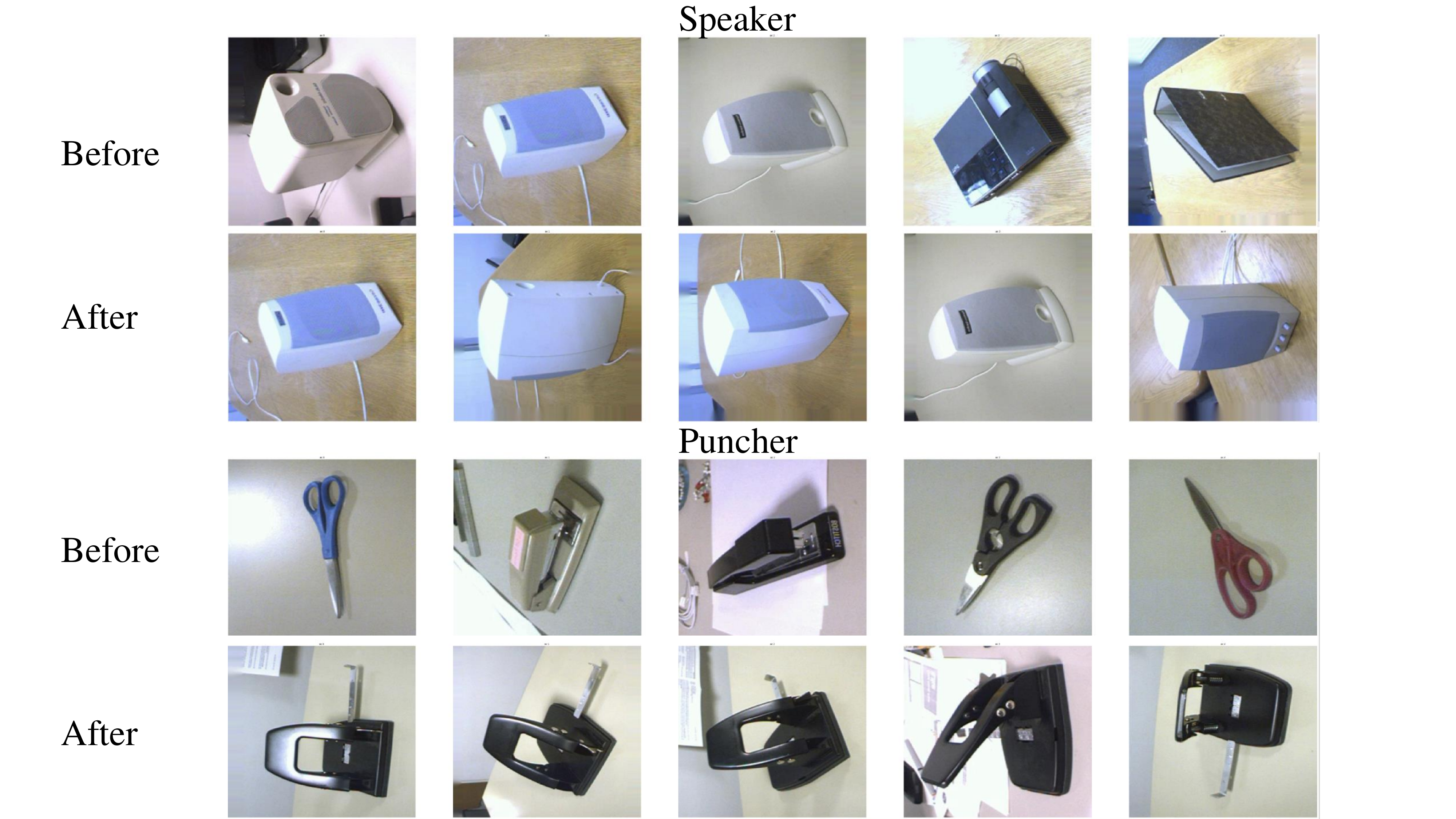}%
        \caption{\small K-nearest neighbors of the prototypes of two different categories in Office-31 - Webcam: ``Speaker'' and ``Puncher''. For each category, the top row exhibits the top-5 neighbors of the prototype of the CLIP model before applying \ours, whereas the bottom row shows the top-5 neighbors of the prototype of the CLIP model after applying \ours. 
       } \vspace{-2mm}
        \label{fig:proto_visualization}%
\end{figure}

\begin{figure}[t!]

    \centering
    \includegraphics[width=0.45\textwidth]{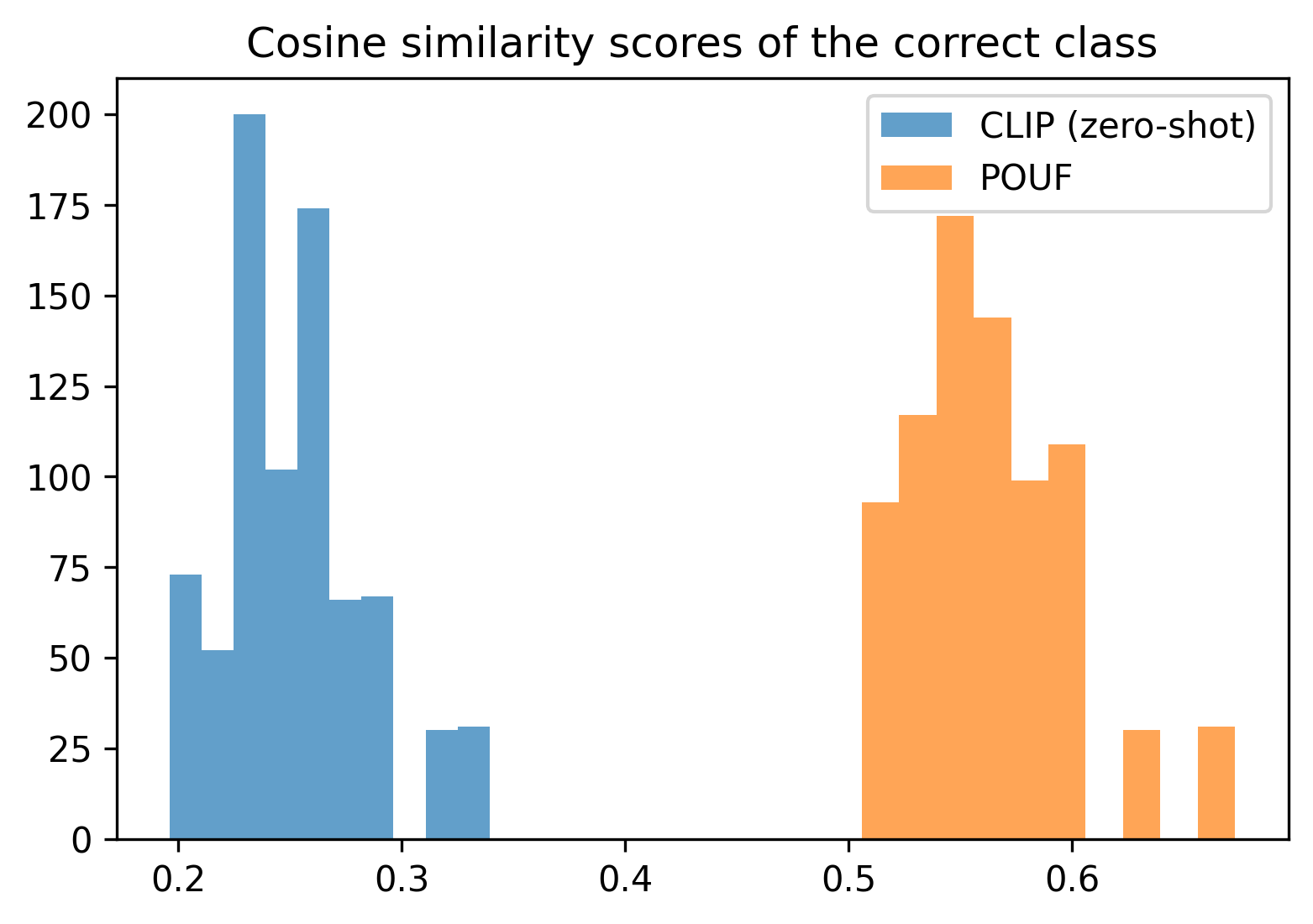}%
    \vspace{-2mm}
        \caption{\small Histograms of cosine similarity scores between the correct classes' prototypes and target features in Office-31 - Webcam.
       }  \vspace{-2mm}    
       \label{fig:cosine_hist}%
\end{figure}
    
    \textbf{Visualization.}
    \textbf{Q: Does \ours\ learn more meaningful prototypes?\ A: Yes} In Figure \ref{fig:proto_visualization}, we visualize the $K$-nearest neighbors ($K=5$) of the prototypes to understand what the model learns. Before applying \ours, the prototypes of the zero-shot CLIP model contain images that do not semantically represent the classes. 
    The fourth and fifth images among the nearest neighbors of the ``Speaker'' prototype include a picture of a projector and a ring binder, respectively,
    while the five nearest neighbors of the ``Puncher'' prototype do not have a single image of a puncher. The ``Puncher'' prototype seems to be close to the images of Scissors and Staplers. After applying \ours, we observe significant improvement qualitatively. Both the ``Speaker'' and ``Puncher'' prototypes are now close to the images of their respective classes, demonstrating that \ours\ helps learn more meaningful prototypes. \textbf{Q: Does \ours\ reduce the violation of the cluster assumption? A: Yes.}
In Figure~\ref{fig:tsne}, we visualize the t-SNE plots of the prototypes and target features. After applying \ours, the prototypes are assigned to their respective clusters of target features. To further quantify this point, for each example, we compute the cosine similarity between the target feature and the prototype of its label. We then visualize the histograms of the similarity scores in Figure \ref{fig:cosine_hist}. It is evident that the cosine similarity scores of \ours\ are higher than those of the zero-shot CLIP model, meaning that the target examples are moved closer to the prototypes of their corresponding classes.

\section{Conclusion}
In this paper, we present \ours, a simple yet effective framework for directly adapting prompt-based zero-shot models to the unlabeled target data. We propose aligning the prototypes and target data in the latent space through transport-based distribution alignment and mutual information maximization. Across 13 image and 15 language tasks, \ours\ achieves consistent performance gains over the baselines. 

\section*{Acknowledgments}
We thank Camillia Smith Barnes for providing valuable feedback on the paper. 
K. Tanwisuth, S. Zhang, H. Zheng, and M. Zhou acknowledge the support of NSF-IIS 2212418 and
Texas Advanced Computing Center.

\bibliographystyle{icml2023}
\bibliography{reference.bib}

\begin{thebibliography}{67}
\providecommand{\natexlab}[1]{#1}
\providecommand{\url}[1]{\texttt{#1}}
\expandafter\ifx\csname urlstyle\endcsname\relax
  \providecommand{\doi}[1]{doi: #1}\else
  \providecommand{\doi}{doi: \begingroup \urlstyle{rm}\Url}\fi

\bibitem[Asano et~al.(2019)Asano, Rupprecht, and Vedaldi]{asano2019self}
Asano, Y.~M., Rupprecht, C., and Vedaldi, A.
\newblock Self-labelling via simultaneous clustering and representation
  learning.
\newblock \emph{arXiv preprint arXiv:1911.05371}, 2019.

\bibitem[Brown et~al.(2020)Brown, Mann, Ryder, Subbiah, Kaplan, Dhariwal,
  Neelakantan, Shyam, Sastry, Askell, Agarwal, Herbert-Voss, Krueger, Henighan,
  Child, Ramesh, Ziegler, Wu, Winter, Hesse, Chen, Sigler, Litwin, Gray, Chess,
  Clark, Berner, McCandlish, Radford, Sutskever, and Amodei]{brown2020language}
Brown, T.~B., Mann, B., Ryder, N., Subbiah, M., Kaplan, J., Dhariwal, P.,
  Neelakantan, A., Shyam, P., Sastry, G., Askell, A., Agarwal, S.,
  Herbert-Voss, A., Krueger, G., Henighan, T., Child, R., Ramesh, A., Ziegler,
  D.~M., Wu, J., Winter, C., Hesse, C., Chen, M., Sigler, E., Litwin, M., Gray,
  S., Chess, B., Clark, J., Berner, C., McCandlish, S., Radford, A., Sutskever,
  I., and Amodei, D.
\newblock Language models are few-shot learners.
\newblock In \emph{NeurIPS}, 2020.

\bibitem[Cao et~al.(2018)Cao, Long, Wang, and Jordan]{cao2018partial}
Cao, Z., Long, M., Wang, J., and Jordan, M.~I.
\newblock Partial transfer learning with selective adversarial networks.
\newblock In \emph{Proceedings of the IEEE conference on computer vision and
  pattern recognition}, pp.\  2724--2732, 2018.

\bibitem[Caron et~al.(2020)Caron, Misra, Mairal, Goyal, Bojanowski, and
  Joulin]{caron2020unsupervised}
Caron, M., Misra, I., Mairal, J., Goyal, P., Bojanowski, P., and Joulin, A.
\newblock Unsupervised learning of visual features by contrasting cluster
  assignments.
\newblock \emph{Advances in Neural Information Processing Systems},
  33:\penalty0 9912--9924, 2020.

\bibitem[Chizat et~al.(2018)Chizat, Peyr{\'e}, Schmitzer, and
  Vialard]{chizat2018unbalanced}
Chizat, L., Peyr{\'e}, G., Schmitzer, B., and Vialard, F.-X.
\newblock Unbalanced optimal transport: Dynamic and kantorovich formulations.
\newblock \emph{Journal of Functional Analysis}, 274\penalty0 (11):\penalty0
  3090--3123, 2018.

\bibitem[Cuturi(2013)]{cuturi2013sinkhorn}
Cuturi, M.
\newblock Sinkhorn distances: Lightspeed computation of optimal transport.
\newblock \emph{Advances in neural information processing systems},
  26:\penalty0 2292--2300, 2013.

\bibitem[Devlin et~al.(2018)Devlin, Chang, Lee, and Toutanova]{devlin2018bert}
Devlin, J., Chang, M.-W., Lee, K., and Toutanova, K.
\newblock Bert: Pre-training of deep bidirectional transformers for language
  understanding.
\newblock \emph{arXiv preprint arXiv:1810.04805}, 2018.

\bibitem[Du et~al.(2022)Du, Liu, Li, and Zhao]{du2022survey}
Du, Y., Liu, Z., Li, J., and Zhao, W.~X.
\newblock A survey of vision-language pre-trained models.
\newblock \emph{arXiv preprint arXiv:2202.10936}, 2022.

\bibitem[Fan et~al.(2020)Fan, Zhang, Chen, and Zhou]{fan2020bayesian}
Fan, X., Zhang, S., Chen, B., and Zhou, M.
\newblock Bayesian attention modules.
\newblock \emph{Advances in Neural Information Processing Systems},
  33:\penalty0 16362--16376, 2020.

\bibitem[Ganin \& Lempitsky(2015)Ganin and Lempitsky]{ganin2015dann}
Ganin, Y. and Lempitsky, V.~S.
\newblock Unsupervised domain adaptation by backpropagation.
\newblock In Bach, F.~R. and Blei, D.~M. (eds.), \emph{ICML}, volume~37 of
  \emph{JMLR Workshop and Conference Proceedings}, pp.\  1180--1189. JMLR.org,
  2015.

\bibitem[Gao et~al.(2020)Gao, Fisch, and Chen]{gao2020making}
Gao, T., Fisch, A., and Chen, D.
\newblock Making pre-trained language models better few-shot learners.
\newblock \emph{arXiv preprint arXiv:2012.15723}, 2020.

\bibitem[Grandvalet et~al.(2005)Grandvalet, Bengio, et~al.]{grandvalet2005semi}
Grandvalet, Y., Bengio, Y., et~al.
\newblock Semi-supervised learning by entropy minimization.
\newblock In \emph{CAP}, pp.\  281--296, 2005.

\bibitem[Guo et~al.(2022)Guo, Tian, Zhang, Zhou, and Zha]{guo2022learning}
Guo, D., Tian, L., Zhang, M., Zhou, M., and Zha, H.
\newblock Learning prototype-oriented set representations for meta-learning.
\newblock In \emph{International Conference on Learning Representations}, 2022.
\newblock URL \url{https://openreview.net/forum?id=WH6u2SvlLp4}.

\bibitem[Huang et~al.(2022)Huang, Chu, and Wei]{huang2022unsupervised}
Huang, T., Chu, J., and Wei, F.
\newblock Unsupervised prompt learning for vision-language models.
\newblock \emph{arXiv preprint arXiv:2204.03649}, 2022.

\bibitem[Jia et~al.(2021)Jia, Yang, Xia, Chen, Parekh, Pham, Le, Sung, Li, and
  Duerig]{jia2021scaling}
Jia, C., Yang, Y., Xia, Y., Chen, Y.-T., Parekh, Z., Pham, H., Le, Q., Sung,
  Y.-H., Li, Z., and Duerig, T.
\newblock Scaling up visual and vision-language representation learning with
  noisy text supervision.
\newblock In \emph{International Conference on Machine Learning}, pp.\
  4904--4916. PMLR, 2021.

\bibitem[Jiang et~al.(2022)Jiang, Shu, Wang, and
  Long]{jiang2022transferability}
Jiang, J., Shu, Y., Wang, J., and Long, M.
\newblock Transferability in deep learning: A survey, 2022.

\bibitem[Kang et~al.(2019)Kang, Jiang, Yang, and
  Hauptmann]{kang2019contrastive}
Kang, G., Jiang, L., Yang, Y., and Hauptmann, A.~G.
\newblock Contrastive adaptation network for unsupervised domain adaptation.
\newblock In \emph{Proceedings of the IEEE/CVF Conference on Computer Vision
  and Pattern Recognition}, pp.\  4893--4902, 2019.

\bibitem[Krause et~al.(2010)Krause, Perona, and
  Gomes]{krause2010discriminative}
Krause, A., Perona, P., and Gomes, R.
\newblock Discriminative clustering by regularized information maximization.
\newblock \emph{Advances in neural information processing systems}, 23, 2010.

\bibitem[Kundu et~al.(2020{\natexlab{a}})Kundu, Venkat, Babu,
  et~al.]{kundu2020universal}
Kundu, J.~N., Venkat, N., Babu, R.~V., et~al.
\newblock Universal source-free domain adaptation.
\newblock In \emph{Proceedings of the IEEE/CVF Conference on Computer Vision
  and Pattern Recognition}, pp.\  4544--4553, 2020{\natexlab{a}}.

\bibitem[Kundu et~al.(2020{\natexlab{b}})Kundu, Venkatesh, Venkat, Revanur, and
  Babu]{kundu2020class}
Kundu, J.~N., Venkatesh, R.~M., Venkat, N., Revanur, A., and Babu, R.~V.
\newblock Class-incremental domain adaptation.
\newblock In \emph{Computer Vision--ECCV 2020: 16th European Conference,
  Glasgow, UK, August 23--28, 2020, Proceedings, Part XIII 16}, pp.\  53--69.
  Springer, 2020{\natexlab{b}}.

\bibitem[Kurmi et~al.(2021)Kurmi, Subramanian, and Namboodiri]{kurmi2021domain}
Kurmi, V.~K., Subramanian, V.~K., and Namboodiri, V.~P.
\newblock Domain impression: A source data free domain adaptation method.
\newblock In \emph{Proceedings of the IEEE/CVF Winter Conference on
  Applications of Computer Vision}, pp.\  615--625, 2021.

\bibitem[Lester et~al.(2021)Lester, Al-Rfou, and Constant]{lester2021power}
Lester, B., Al-Rfou, R., and Constant, N.
\newblock The power of scale for parameter-efficient prompt tuning.
\newblock \emph{arXiv preprint arXiv:2104.08691}, 2021.

\bibitem[Li et~al.(2020{\natexlab{a}})Li, Zhou, Xiong, and
  Hoi]{li2020prototypical}
Li, J., Zhou, P., Xiong, C., and Hoi, S.~C.
\newblock Prototypical contrastive learning of unsupervised representations.
\newblock \emph{arXiv preprint arXiv:2005.04966}, 2020{\natexlab{a}}.

\bibitem[Li et~al.(2020{\natexlab{b}})Li, Jiao, Cao, Wong, and Wu]{li2020model}
Li, R., Jiao, Q., Cao, W., Wong, H.-S., and Wu, S.
\newblock Model adaptation: Unsupervised domain adaptation without source data.
\newblock In \emph{Proceedings of the IEEE/CVF Conference on Computer Vision
  and Pattern Recognition}, pp.\  9641--9650, 2020{\natexlab{b}}.

\bibitem[Li \& Liang(2021)Li and Liang]{li2021prefix}
Li, X.~L. and Liang, P.
\newblock Prefix-tuning: Optimizing continuous prompts for generation.
\newblock \emph{arXiv preprint arXiv:2101.00190}, 2021.

\bibitem[Liang et~al.(2020)Liang, Hu, and Feng]{liang2020we}
Liang, J., Hu, D., and Feng, J.
\newblock Do we really need to access the source data? source hypothesis
  transfer for unsupervised domain adaptation.
\newblock In \emph{International Conference on Machine Learning}, pp.\
  6028--6039. PMLR, 2020.

\bibitem[Liu et~al.(2021{\natexlab{a}})Liu, Yuan, Fu, Jiang, Hayashi, and
  Neubig]{liu2021pre}
Liu, P., Yuan, W., Fu, J., Jiang, Z., Hayashi, H., and Neubig, G.
\newblock Pre-train, prompt, and predict: A systematic survey of prompting
  methods in natural language processing.
\newblock \emph{arXiv preprint arXiv:2107.13586}, 2021{\natexlab{a}}.

\bibitem[Liu et~al.(2021{\natexlab{b}})Liu, Gong, Wu, Zhang, Su, and
  Liu]{liu2021fusedream}
Liu, X., Gong, C., Wu, L., Zhang, S., Su, H., and Liu, Q.
\newblock Fusedream: Training-free text-to-image generation with improved clip+
  gan space optimization.
\newblock \emph{arXiv preprint arXiv:2112.01573}, 2021{\natexlab{b}}.

\bibitem[Liu et~al.(2019)Liu, Ott, Goyal, Du, Joshi, Chen, Levy, Lewis,
  Zettlemoyer, and Stoyanov]{liu2019roberta}
Liu, Y., Ott, M., Goyal, N., Du, J., Joshi, M., Chen, D., Levy, O., Lewis, M.,
  Zettlemoyer, L., and Stoyanov, V.
\newblock Roberta: A robustly optimized bert pretraining approach.
\newblock \emph{arXiv preprint arXiv:1907.11692}, 2019.

\bibitem[Long et~al.(2015)Long, Cao, Wang, and Jordan]{long2015dan}
Long, M., Cao, Y., Wang, J., and Jordan, M.~I.
\newblock Learning transferable features with deep adaptation networks.
\newblock In \emph{Proceedings of the 32nd International Conference on Machine
  Learning, {ICML} 2015, Lille, France, 6-11 July 2015}, pp.\  97--105, 2015.

\bibitem[Long et~al.(2017)Long, Zhu, Wang, and Jordan]{long2017jan}
Long, M., Zhu, H., Wang, J., and Jordan, M.~I.
\newblock Deep transfer learning with joint adaptation networks.
\newblock In Precup, D. and Teh, Y.~W. (eds.), \emph{Proceedings of the 34th
  International Conference on Machine Learning}, volume~70 of \emph{Proceedings
  of Machine Learning Research}, pp.\  2208--2217. PMLR, 06--11 Aug 2017.

\bibitem[Long et~al.(2018)Long, Cao, Wang, and Jordan]{long2018cdan}
Long, M., Cao, Z., Wang, J., and Jordan, M.~I.
\newblock Conditional adversarial domain adaptation.
\newblock In Bengio, S., Wallach, H., Larochelle, H., Grauman, K.,
  Cesa-Bianchi, N., and Garnett, R. (eds.), \emph{Advances in Neural
  Information Processing Systems}, volume~31. Curran Associates, Inc., 2018.

\bibitem[M{\'e}rigot \& Oudet(2016)M{\'e}rigot and Oudet]{merigot2016discrete}
M{\'e}rigot, Q. and Oudet, E.
\newblock Discrete optimal transport: complexity, geometry and applications.
\newblock \emph{Discrete \& Computational Geometry}, 55\penalty0 (2):\penalty0
  263--283, 2016.

\bibitem[Morerio et~al.(2017)Morerio, Cavazza, and Murino]{morerio2017minimal}
Morerio, P., Cavazza, J., and Murino, V.
\newblock Minimal-entropy correlation alignment for unsupervised deep domain
  adaptation.
\newblock \emph{arXiv preprint arXiv:1711.10288}, 2017.

\bibitem[Pan et~al.(2019)Pan, Yao, Li, Wang, Ngo, and
  Mei]{pan2019transferrable}
Pan, Y., Yao, T., Li, Y., Wang, Y., Ngo, C.-W., and Mei, T.
\newblock Transferrable prototypical networks for unsupervised domain
  adaptation.
\newblock In \emph{Proceedings of the IEEE/CVF Conference on Computer Vision
  and Pattern Recognition (CVPR)}, 2019.

\bibitem[Peng et~al.(2019)Peng, Bai, Xia, Huang, Saenko, and
  Wang]{peng2019moment}
Peng, X., Bai, Q., Xia, X., Huang, Z., Saenko, K., and Wang, B.
\newblock Moment matching for multi-source domain adaptation.
\newblock In \emph{Proceedings of the IEEE/CVF International Conference on
  Computer Vision}, pp.\  1406--1415, 2019.

\bibitem[Qiu et~al.(2020)Qiu, Sun, Xu, Shao, Dai, and Huang]{qiu2020pre}
Qiu, X., Sun, T., Xu, Y., Shao, Y., Dai, N., and Huang, X.
\newblock Pre-trained models for natural language processing: A survey.
\newblock \emph{Science China Technological Sciences}, 63\penalty0
  (10):\penalty0 1872--1897, 2020.

\bibitem[Radford et~al.(2021)Radford, Kim, Hallacy, Ramesh, Goh, Agarwal,
  Sastry, Askell, Mishkin, Clark, et~al.]{radford2021learning}
Radford, A., Kim, J.~W., Hallacy, C., Ramesh, A., Goh, G., Agarwal, S., Sastry,
  G., Askell, A., Mishkin, P., Clark, J., et~al.
\newblock Learning transferable visual models from natural language
  supervision.
\newblock In \emph{International Conference on Machine Learning}, pp.\
  8748--8763. PMLR, 2021.

\bibitem[Saenko et~al.(2010)Saenko, Kulis, Fritz, and
  Darrell]{saenko2010adapting}
Saenko, K., Kulis, B., Fritz, M., and Darrell, T.
\newblock Adapting visual category models to new domains.
\newblock In \emph{European conference on computer vision}, pp.\  213--226.
  Springer, 2010.

\bibitem[Saito et~al.(2019)Saito, Kim, Sclaroff, Darrell, and
  Saenko]{saito2019semi}
Saito, K., Kim, D., Sclaroff, S., Darrell, T., and Saenko, K.
\newblock Semi-supervised domain adaptation via minimax entropy.
\newblock In \emph{Proceedings of the IEEE/CVF International Conference on
  Computer Vision}, pp.\  8050--8058, 2019.

\bibitem[Saito et~al.(2020)Saito, Kim, Sclaroff, and
  Saenko]{saito2020universal}
Saito, K., Kim, D., Sclaroff, S., and Saenko, K.
\newblock Universal domain adaptation through self supervision.
\newblock \emph{arXiv preprint arXiv:2002.07953}, 2020.

\bibitem[Schick \& Sch{\"u}tze(2020)Schick and
  Sch{\"u}tze]{schick2020exploiting}
Schick, T. and Sch{\"u}tze, H.
\newblock Exploiting cloze questions for few shot text classification and
  natural language inference.
\newblock \emph{arXiv preprint arXiv:2001.07676}, 2020.

\bibitem[Shi \& Sha(2012)Shi and Sha]{shi2012information}
Shi, Y. and Sha, F.
\newblock Information-theoretical learning of discriminative clusters for
  unsupervised domain adaptation.
\newblock \emph{arXiv preprint arXiv:1206.6438}, 2012.

\bibitem[Shin et~al.(2020)Shin, Razeghi, Logan~IV, Wallace, and
  Singh]{shin2020autoprompt}
Shin, T., Razeghi, Y., Logan~IV, R.~L., Wallace, E., and Singh, S.
\newblock Autoprompt: Eliciting knowledge from language models with
  automatically generated prompts.
\newblock \emph{arXiv preprint arXiv:2010.15980}, 2020.

\bibitem[Snell et~al.(2017)Snell, Swersky, and Zemel]{snell2017prototypical}
Snell, J., Swersky, K., and Zemel, R.
\newblock Prototypical networks for few-shot learning.
\newblock \emph{Advances in neural information processing systems}, 30, 2017.

\bibitem[Tanwisuth et~al.(2021)Tanwisuth, Fan, Zheng, Zhang, Zhang, Chen, and
  Zhou]{tanwisuth2021prototype}
Tanwisuth, K., Fan, X., Zheng, H., Zhang, S., Zhang, H., Chen, B., and Zhou, M.
\newblock A prototype-oriented framework for unsupervised domain adaptation.
\newblock \emph{Advances in Neural Information Processing Systems},
  34:\penalty0 17194--17208, 2021.

\bibitem[Tanwisuth et~al.(2023)Tanwisuth, Zhang, He, and
  Zhou]{tanwisuth2023prototype}
Tanwisuth, K., Zhang, S., He, P., and Zhou, M.
\newblock A prototype-oriented clustering for domain shift with source privacy.
\newblock \emph{arXiv preprint arXiv:2302.03807}, 2023.

\bibitem[Tzeng et~al.(2014)Tzeng, Hoffman, Zhang, Saenko, and
  Darrell]{tzeng2014deep}
Tzeng, E., Hoffman, J., Zhang, N., Saenko, K., and Darrell, T.
\newblock Deep domain confusion: Maximizing for domain invariance.
\newblock \emph{arXiv preprint arXiv:1412.3474}, 2014.

\bibitem[Tzeng et~al.(2017)Tzeng, Hoffman, Saenko, and
  Darrell]{tzeng2017adversarial}
Tzeng, E., Hoffman, J., Saenko, K., and Darrell, T.
\newblock Adversarial discriminative domain adaptation.
\newblock In \emph{Proceedings of the IEEE conference on computer vision and
  pattern recognition}, pp.\  7167--7176, 2017.

\bibitem[Venkateswara et~al.(2017)Venkateswara, Eusebio, Chakraborty, and
  Panchanathan]{venkateswara2017deep}
Venkateswara, H., Eusebio, J., Chakraborty, S., and Panchanathan, S.
\newblock Deep hashing network for unsupervised domain adaptation.
\newblock In \emph{Proceedings of the IEEE conference on computer vision and
  pattern recognition}, pp.\  5018--5027, 2017.

\bibitem[Villani(2008)]{villani2008optimal}
Villani, C.
\newblock \emph{Optimal transport: old and new}, volume 338.
\newblock Springer Science \& Business Media, 2008.

\bibitem[Vu et~al.(2019)Vu, Jain, Bucher, Cord, and P{\'e}rez]{vu2019advent}
Vu, T.-H., Jain, H., Bucher, M., Cord, M., and P{\'e}rez, P.
\newblock Advent: Adversarial entropy minimization for domain adaptation in
  semantic segmentation.
\newblock In \emph{Proceedings of the IEEE/CVF Conference on Computer Vision
  and Pattern Recognition}, pp.\  2517--2526, 2019.

\bibitem[Wallace et~al.(2019)Wallace, Feng, Kandpal, Gardner, and
  Singh]{wallace2019universal}
Wallace, E., Feng, S., Kandpal, N., Gardner, M., and Singh, S.
\newblock Universal adversarial triggers for attacking and analyzing nlp.
\newblock \emph{arXiv preprint arXiv:1908.07125}, 2019.

\bibitem[Wang et~al.(2018)Wang, Singh, Michael, Hill, Levy, and
  Bowman]{wang2018glue}
Wang, A., Singh, A., Michael, J., Hill, F., Levy, O., and Bowman, S.~R.
\newblock Glue: A multi-task benchmark and analysis platform for natural
  language understanding.
\newblock \emph{arXiv preprint arXiv:1804.07461}, 2018.

\bibitem[Wang et~al.(2020)Wang, Shelhamer, Liu, Olshausen, and
  Darrell]{wang2020tent}
Wang, D., Shelhamer, E., Liu, S., Olshausen, B., and Darrell, T.
\newblock Tent: Fully test-time adaptation by entropy minimization.
\newblock \emph{arXiv preprint arXiv:2006.10726}, 2020.

\bibitem[Wang et~al.(2022)Wang, Guo, Zhao, Zheng, Tanwisuth, Chen, and
  Zhou]{wang2022representing}
Wang, D., Guo, D., Zhao, H., Zheng, H., Tanwisuth, K., Chen, B., and Zhou, M.
\newblock Representing mixtures of word embeddings with mixtures of topic
  embeddings.
\newblock In \emph{International Conference on Learning Representations}, 2022.
\newblock URL \url{https://openreview.net/forum?id=IYMuTbGzjFU}.

\bibitem[Wenzel et~al.(2022)Wenzel, Dittadi, Gehler, Simon-Gabriel, Horn,
  Zietlow, Kernert, Russell, Brox, Schiele, et~al.]{wenzel2022assaying}
Wenzel, F., Dittadi, A., Gehler, P.~V., Simon-Gabriel, C.-J., Horn, M.,
  Zietlow, D., Kernert, D., Russell, C., Brox, T., Schiele, B., et~al.
\newblock Assaying out-of-distribution generalization in transfer learning.
\newblock \emph{arXiv preprint arXiv:2207.09239}, 2022.

\bibitem[Wu et~al.(2020)Wu, Zhou, Yang, Zhao, Latecki, et~al.]{wu2020entropy}
Wu, X., Zhou, Q., Yang, Z., Zhao, C., Latecki, L.~J., et~al.
\newblock Entropy minimization vs. diversity maximization for domain
  adaptation.
\newblock \emph{arXiv preprint arXiv:2002.01690}, 2020.

\bibitem[Yue et~al.(2021)Yue, Zheng, Zhang, Gao, Darrell, Keutzer, and
  Vincentelli]{yue2021prototypical}
Yue, X., Zheng, Z., Zhang, S., Gao, Y., Darrell, T., Keutzer, K., and
  Vincentelli, A.~S.
\newblock Prototypical cross-domain self-supervised learning for few-shot
  unsupervised domain adaptation.
\newblock In \emph{Proceedings of the IEEE/CVF Conference on Computer Vision
  and Pattern Recognition}, pp.\  13834--13844, 2021.

\bibitem[Zhang et~al.(2021{\natexlab{a}})Zhang, Fan, Chen, and
  Zhou]{zhang2021bayesian}
Zhang, S., Fan, X., Chen, B., and Zhou, M.
\newblock Bayesian attention belief networks.
\newblock In \emph{International Conference on Machine Learning}, pp.\
  12413--12426. PMLR, 2021{\natexlab{a}}.

\bibitem[Zhang et~al.(2021{\natexlab{b}})Zhang, Fan, Zheng, Tanwisuth, and
  Zhou]{zhang2021alignment}
Zhang, S., Fan, X., Zheng, H., Tanwisuth, K., and Zhou, M.
\newblock Alignment attention by matching key and query distributions.
\newblock In \emph{Neural Information Processing Systems}, Dec.
  2021{\natexlab{b}}.

\bibitem[Zhang et~al.(2021{\natexlab{c}})Zhang, Gong, and
  Choi]{zhang2021capturing}
Zhang, S., Gong, C., and Choi, E.
\newblock Capturing label distribution: A case study in nli.
\newblock \emph{arXiv preprint arXiv:2102.06859}, 2021{\natexlab{c}}.

\bibitem[Zhang et~al.(2021{\natexlab{d}})Zhang, Gong, and
  Choi]{zhang2021learning}
Zhang, S., Gong, C., and Choi, E.
\newblock Learning with different amounts of annotation: From zero to many
  labels.
\newblock \emph{arXiv preprint arXiv:2109.04408}, 2021{\natexlab{d}}.

\bibitem[Zhang et~al.(2022)Zhang, Gong, Liu, He, Chen, and
  Zhou]{zhang2022allsh}
Zhang, S., Gong, C., Liu, X., He, P., Chen, W., and Zhou, M.
\newblock Allsh: Active learning guided by local sensitivity and hardness.
\newblock \emph{arXiv preprint arXiv:2205.04980}, 2022.

\bibitem[Zheng \& Zhou(2021)Zheng and Zhou]{zheng2021exploiting}
Zheng, H. and Zhou, M.
\newblock Exploiting chain rule and {B}ayes' theorem to compare probability
  distributions.
\newblock \emph{Advances in Neural Information Processing Systems},
  34:\penalty0 14993--15006, 2021.

\bibitem[Zhou et~al.(2022{\natexlab{a}})Zhou, Yang, Loy, and
  Liu]{zhou2022conditional}
Zhou, K., Yang, J., Loy, C.~C., and Liu, Z.
\newblock Conditional prompt learning for vision-language models.
\newblock In \emph{Proceedings of the IEEE/CVF Conference on Computer Vision
  and Pattern Recognition}, pp.\  16816--16825, 2022{\natexlab{a}}.

\bibitem[Zhou et~al.(2022{\natexlab{b}})Zhou, Yang, Loy, and
  Liu]{zhou2022learning}
Zhou, K., Yang, J., Loy, C.~C., and Liu, Z.
\newblock Learning to prompt for vision-language models.
\newblock \emph{International Journal of Computer Vision}, 130\penalty0
  (9):\penalty0 2337--2348, 2022{\natexlab{b}}.

\end{thebibliography}


\newpage
\appendix
\onecolumn

\begin{center}
    {\textbf{\LARGE POUF: Prompt-oriented unsupervised fine-tuning \\ \vspace{2mm}for large pre-trained models: Appendix}}\\
    \vspace{4mm}
\end{center}

\section{Detailed implementation details}
\label{appendix:implementation}
\subsection{Language-augmented Vision models}

We build our method using the open-source CLIP codebase \citep{radford2021learning, liu2021fusedream} and TLlib transfer learning library \citep{jiang2022transferability}. For all experiments, we adopt the ViTB-16 for the image encoder and the default transformer from the CLIP paper for the text encoder. All the unlabeled target samples are used for fine-tuning. The learning rate shcedule is set to $\eta_{\text{iter}}=\eta_0(1+\gamma\text{iter})^{-\alpha}$, where $\eta_0$ is the initial learning rate. We adopt the following default hyper-parameters: $\gamma=0.0002$, and $\alpha=0.75$. We set $\eta_0=5\mathrm{e}{-7}$ for all experiments except for prompt tuning on Office-31 where $\eta_0=1\mathrm{e}{-3}$. For prompt tuning, the context length, $p$, is set to $4$. We then initialize the soft-prompt parameters with the embeddings of the words ``A photo of a''. We use a mini-batch SGD with a momentum of $0.9$ and a batch size of $96$ for Office-31 and Office-Home and $16$ for DomainNet. The weight of the mutual-information objective, $\lambda$, is set to $0.3$ for all experiments. We run all experiments for $5,000$ iterations using the seeds $\{0, 1, 2\}$ and report the average accuracy. All experiments are conducted using a single Nvidia Tesla V100 GPU.

For all the baselines, we use the same hyper-parameters as our method for prompt tuning. For Tent, we set the weight of the entropy objective to $0.3$, the same as the weight of the mutual-information objective, $\lambda$. For UPL, we follow the original paper to set the number of pseudo labels as $16$ examples per class. For CoOp, we set the number of few-shot examples per class as $16$ per the original paper.

\subsection{Masked-language models}
We follow the experimental protocols in \citet{gao2020making}. Specifically, for each task, the data is split into $\mathcal{D}_{train}$, $\mathcal{D}_{dev}$, and $\mathcal{D}_{test}$. The authors tune the hyper-parameters on $\mathcal{D}_{dev}$ and report the performance of the model on $\mathcal{D}_{test}$. We take the same hyper-parameters as the original paper: batch size $=8$, learning rate $=1e-5$, training steps $=1,000$, and few-shot examples per class $=16$. The weight of the mutual-information objective, $\lambda$, is set to $0.6$ for all experiments except for the unsupervised setting on the MRPC and QQP tasks. In these two tasks, the weights are set to $1\mathrm{e}{-4}$ for the mutual-information objective and to $1\mathrm{e}{-2}$ for the transport cost. We validate the performance of the model every $100$ steps on the development set and take the best validated checkpoint for the final evaluation on the test set. For the unsupervised setting, we use the dataset provided by \citet{gao2020making} without the labels. For the few-shot setting, we incorporate POUF into LM-BFF by using the few-shot examples without labels to compute POUF's loss and add it to LM-BFF's objective. We run all experiments using the seeds $\{42, 21, 87, 13, 100\}$ and report the average accuracy. All experiments are conducted on four Nvidia Tesla V100 GPUs.

\section{Parameter size and runtime analysis}
\label{appendix:analysis}

\begin{table}[h] 
\centering
\caption{
Parameter and run-time analyses of prompt v.s. model tuning the CLIP model.}
\vspace{-2mm}
{\begin{tabular}{c|c|c|c}
\toprule  Settings & \multicolumn{2}{c|}{Parameter analysis} & \multicolumn{1}{c}{Run-time analysis}   \\
\cmidrule(r){2-3} \cmidrule(r){4-4} 
\multicolumn{1}{c}{} & \multicolumn{1}{|c|}{ Tunable parameters} & \multicolumn{1}{c|}{ Total parameters} & \multicolumn{1}{c}{ Time per iteration (s)}  \\
\midrule
Prompt tuning & $2,049$ &  $149,622,785$ & $0.4$ \\
Model tuning & $149,620,737$ & $149,620,737$ & $1.8$\\

\bottomrule
\end{tabular}}
\label{tab:parameter}
\end{table}

In Table \ref{tab:parameter}, we present parameter and run-time analyses of prompt and model tuning for the CLIP model. Prompt tuning introduces $h\ (\text{Embedding dimension}) \times p\ (\text{Context length}) = 512 \times 4 = 2048$ number of parameters to the model, increasing the total number of parameters slightly. However, given the size of the CLIP model, the increase is negligible. We note that the number of tunable parameters is $2,049$ since we also adjust the temperature parameter, a scalar, in the CLIP model. Since the number of tunable parameters is significantly lower in prompt tuning than in model tuning, it is not surprising that the training time per iteration for prompt tuning is one-quarter that of model tuning.

\newpage
\section{Limitation and societal impact}
\label{appendix:broader_impact}
\ours\ leverages the power of language representations to adapt to the unlabeled target data directly. While the language model may have a large vocabulary size, it is possible that it does not capture some exotic words. This means that if the categories that we want to predict are outside the vocabulary of the language model, we cannot use our method to fine-tune the model, as we cannot even make predictions. Future research can focus on how to extend \ours\ to adapt to unknown classes beyond the vocabulary of the model.

\ours\ relies on large-scale pre-trained models to adapt to the target data. As with any computationally intensive venture, practitioners should consider using sustainable energy resources. On the positive side, \ours\ enables practitioners to efficiently fine-tune on the target data, saving the computation.

\section{Datasets}
\label{appendix:datasets}

\begin{table}[htp!]
\centering
\caption{Dataset information for image experiments.}
 \resizebox{1\textwidth}{!}{\begin{tabular}{lcccc}
\hline Dataset & Number of images  & Number of categories & Domains \\
\hline 
Office-31 \cite{saenko2010adapting} &$4,652$&  $31$ &\text{Amazon, DSLR, Webcam}\\
Office-Home \cite{venkateswara2017deep}  &$ 15,500$ & $65$ & \text{Artistic images, Clip Art,
Product images, Real-World }  \\
DomainNet \cite{peng2019moment} & $569,010$ & $345$ &  \text{Clipart, Infograph, Painting, Quickdraw, Real, Sketch}\\
\hline
\end{tabular}}
\label{tab:dataset}
\end{table}

\begin{table}[h] 
\centering
\caption{Dataset information for language experiments from \citet{gao2020making}. $|\mathcal{Y}|$ denotes the number of categories. L refers to the average number of words in a sentence. The license of the GLUE benchmark is cc-by-4.0.}
{\begin{tabular}{llrrrrcl}
\hline Category & Dataset & $|\mathcal{Y}|$ & $L$ & \#Train & \#Test & Type & Labels (classification tasks) \\
\hline & SST-2 & $2$ & $19$ & $6,920$ & $872$ & sentiment & positive, negative \\
& SST-5 & $5$ & $18$ & $8,544$ &$ 2,210$ & sentiment & v. pos., positive, neutral, negative, v. neg. \\
& MR & $2$ & $20$ &$ 8,662$ & $2,000$ & sentiment & positive, negative \\
single- & CR & $2$ & $19$ & $1,775$ & $2,000$ & sentiment & positive, negative \\
sentence & MPQA & $2$ & $3$ & $8,606$ & $2,000$ & opinion polarity & positive, negative \\
& Subj & $2$ & $23$ & $8,000$ & $2,000$ & subjectivity & subjective, objective \\
& TREC & $6$ & $10$ & $5,452$ & $500$ & question cls. & abbr., entity, description, human, loc., num. \\
& CoLA & $2$ & $8$ & $8,551$ & $1,042$ & acceptability & grammatical, not grammatical \\
\hline  & QNLI & $2$ & $11 / 30$ & $104,743$ & $5,463$ & NLI & entailment, not entailment \\
 & MNLI & 3 & $22 / 11$ & $392,702$ & $9,815$ & NLI & entailment, neutral, contradiction \\
{sentence- } & RTE & $2$ & $49 / 10$ & $2,490$ & $277$ & NLI & entailment, not entailment \\
pair & MRPC & $2$ & $22 / 21$ & $3,668$ & $408$ & paraphrase & equivalent, not equivalent \\
& QQP & $2$ & $12 / 12$ & $363,846$ & $40,431$ & paraphrase & equivalent, not equivalent \\
& STS-B & $\mathcal{R}$ & $11 / 11$ & $5,749$ & $1,500$ & sent. similarity & - \\
\hline
\end{tabular}}
\end{table}

\section{Unseen Class Experiments}
\label{appendix:unseen}
Testing POUF under an open class scenario is an interesting setting. 
POUF is developed to address domain shift (changes in feature distribution) but does not specifically target label space shift. To investigate the impact of POUF on generalization to unseen classes, we designed two experiments: in-domain generalization and out-of-domain generalization. The setup of each experiment is explained below. POUF (text prototypes = seen classes): we adapt POUF with the textual prototypes from only the seen classes.
POUF (text prototypes = seen + unseen classes): in many cases, the target users may not have image data of the unseen classes but know the class names in advance. In this version, we adapt POUF with the textual prototypes from both the seen and unseen classes. 1) Experiment 1: In-domain generalization. In this experiment, our goal is to investigate the impact of POUF on both seen and unseen image classes within the same domain. To do so, we randomly divide the classes in one of the Office-31 dataset's domains into two groups: half as seen and the other half as unseen. We employ images from the seen classes for adaptation using POUF. To evaluate the model, we assess its performance on both seen and unseen classes within this domain. The findings are presented below. 2) Experiment 2: Out-of-domain generalization. In this experiment, we aim to examine the impact of POUF on seen and unseen image classes across different domains. To achieve this, we randomly split the classes into two groups: half as seen and the other half as unseen. We apply POUF to adapt the model using the seen classes from one domain and then evaluate the adapted model on the seen and unseen classes of another domain. The findings are presented in Tables \ref{tab:unseen_webcam_class_ablation} and \ref{tab:unseen_amazon_class_ablation}.

In both tables, POUF enhances the performance of zero-shot models for seen classes. However, when trained exclusively with seen-class prototypes, POUF yields improved performance for the seen class but slightly weaker generalization for the unseen class when evaluated on the Amazon domain, in both in-domain and out-of-domain generalization settings. This outcome highlights a minor trade-off between the model's specialization and generalization. If target users have access to unseen class names, training POUF with both seen and unseen class text prototypes results in superior performance compared to the zero-shot model for both seen and unseen classes in both scenarios.
The outcome implies that the essential concept is to maintain the semantics in the text space for improved generalization. A possible explanation for this observation is that, during pre-training, both text and image features are plentiful. However, during fine-tuning, we continue to have abundant image features but limited text prompts, which come from the target user. If we fine-tune both text and image encoders, we might end up with representations that are particularly tailored to the seen classes.

\begin{table}[htp!]
\centering
\caption{\small POUF on unseen classes for Webcam (W).}
\vspace{-2mm}
 \resizebox{1.0\textwidth}{!}{ \begin{tabular}{c|cc|cc}
\toprule
& \multicolumn{2}{c|}{In-domain Setting}  & \multicolumn{2}{c}{Out-of-domain Setting}\\
Adaptation & seen class of Webcam (W)	 & seen class of Webcam (W)	& seen class of Webcam (W)	& seen class of Webcam (W) 
 \\ \midrule
 Testing & seen class of W & unseen class of W &	seen class of A & unseen class of A \\
\midrule
CLIP (zero-shot) & $65.2$ & $87.5$ & $76.5$ & $83.0$\\
POUF (text prototypes = seen classes) & $\bf{85.1}$ & $92.8$& $\bf{83.6}$ &$82.7$\\
POUF (text prototypes = seen + unseen classes)	 & $75.5$& $\bf{95.7}$&$79.6$& $\bf{85.8}$\\
\bottomrule
\end{tabular}}\vspace{-2mm}
\label{tab:unseen_webcam_class_ablation}
\end{table}

\begin{table}[htp!]
\centering
\caption{\small POUF on unseen classes for Amazon (A).}
\vspace{-2mm}
 \resizebox{1.0\textwidth}{!}{ \begin{tabular}{c|cc|cc}
\toprule
& \multicolumn{2}{c|}{In-domain Setting}  & \multicolumn{2}{c}{Out-of-domain Setting}\\
Adaptation & seen class of Amazon (A) & seen class of Amazon (A) & seen class of Amazon (A) & seen class of Amazon (A)

 \\ \midrule
 Testing & seen class of A & unseen class of A & seen class of W & unseen class of W\\
\midrule
CLIP (zero-shot) & $76.5$ & $83.0$ & $65.2$ & $87.5$\\
POUF (text prototypes = seen classes) & $\bf{84.1}$& $81.7$& $\bf{81.1}$ & $89.4$\\
POUF (text prototypes = seen + unseen classes)	 & $80.6$ & $\bf{87.2}$& $72.2$& $\bf{93.1}$\\
\bottomrule
\end{tabular}}\vspace{-2mm}
\label{tab:unseen_amazon_class_ablation}
\end{table}

\section{Additional ablation study on the hyper-parameter $\lambda$}
\begin{table}[!ht]
    \centering
    \begin{tabular}{l|l|l|l|l|l|l|l|l|l|l}
        $\lambda$ & 0.1 & 0.2 & 0.3 & 0.4 & 0.5 & 0.6 & 0.7 & 0.8 & 0.9 & 1 \\ \hline
        Accuracy & 87.5 & 88.3 & 90.6 & 91.7 & 90.8 & 89.2 & 90.6 & 89.8 & 88.9 & 91.6 \\ 
    \end{tabular}
    \caption{Accuracy of POUF on the Webcam domain of the Office-31 dataset for different $\lambda$
 values.}
\end{table}

\section{Class proportion estimation}
\label{appendix:proportion_esimation}
The transport framework provides a nice way of handling the class-imbalanced issue. If we know the proportions beforehand, we can adjust the marginal distribution, $p(\wv_k)$, by setting it to the corresponding proportions. Classes with higher proportions will then receive higher transport costs. However, in practice, the proportions are unknown so we need to estimate them. To estimate the proportion of class, we can marginalize the predictive distribution of the classes given the data as follows
$$p(\wv_k) = \sum_i \pi_{\thetav}(\wv_{k}\given  \fv_{i})= \sum_i \frac{ p(\wv_k)\exp(\wv_k^T\fv_i)}{\sum_{k'=1}^Kp(\wv_{k'})\exp(\wv_{k'}^T\fv_i  )}$$
Since the marginalization is done over the data, mini-batch update is often required to overcome the computational expense. Thus, the above estimation can be noisy. To overcome this issue and estimate the global proportions from a local mini-batch of data, we can iteratively learn the global proportions over multiple iterations. We can then learn the proportions as follows

$$\tilde{p}(\wv_k)^{l+1} = \frac{1}{M}\sum_i \pi_{\thetav}(\wv_{k}\given  \fv_{i})$$

$$p(\wv_k)^{l+1} = \alpha ^l\tilde{p}(\wv_k)^{l+1} + (1-\alpha^l) p(\wv_k)^l, $$

where $\alpha^l$ is the learning rate of iteration $l$ and follows a cosine learning rate schedule.

\section{Pseudo code}

\begin{algorithm}[!htp]
\caption{\ours\ Pseudocode for language-augmented vision models, PyTorch-like}
\label{alg:code}
\definecolor{codeblue}{rgb}{0.25,0.5,0.5}
\definecolor{codekw}{rgb}{0.85, 0.18, 0.50}
\lstset{
  backgroundcolor=\color{white},
  basicstyle=\fontsize{7.5pt}{7.5pt}\ttfamily\selectfont,
  columns=fullflexible,
  breaklines=true,
  captionpos=b,
  commentstyle=\fontsize{7.5pt}{7.5pt}\color{codeblue},
  keywordstyle=\fontsize{7.5pt}{7.5pt}\color{codekw},
}
\begin{lstlisting}[language=python]
# F: image encoder
# G: text encoder
# prompts: textual inputs (i.e. "A photo of {CLASS}")

for x in loader:  # Load a minibatch x with M samples
    f, prototypes = F(x), G(prompts) # Compute embeddings
    
    f = normalize(f, dim=1)
    prototypes = normalize(prototypes, dim=1)
    
    sim_mat = f @ prototypes.T  #  M-by-K
    
    L_transport = compute_transport_loss(sim_mat, T)
    L_mi = compute_mi_loss(sim_mat/T)
    L = L_transport + lambda_mi * L_mi # Loss

    L.backward()  # Back-propagate
    update(F, G)  # SGD update 

def compute_transport_loss(sim_mat, T):
    # Compute transport cost
    cost = 1 - (sim_mat)

    # Compute transport plans
    forward_plan = softmax(sim_mat/T, dim=0)
    backward_plan = softmax(sim_mat/T, dim=1)

    # Compute final loss
    forward_cost = (cost * forward_plan).sum(0).mean()
    backward_cost = (cost * backward_plan).sum(1).mean()
    return forward_cost + backward_cost

def compute_mi_loss(sim_mat):
    # Compute conditional entropy
    softmax_out = softmax(sim_mat, dim=1)
    entropy_loss = sum(-softmax_out * log(softmax_out + eps), dim=1).mean()

    # Compute marginal entropy
    mean_softmax = softmax_out.mean(0)
    regularization = sum(-mean_softmax * log(mean_softmax + eps))

    # Compute final loss
    return entropy_loss - regularization
\end{lstlisting}
\end{algorithm}

\begin{algorithm}[htp!]
\caption{\ours\ Pseudocode for masked-language models, PyTorch-like}
\label{alg:code}
\definecolor{codeblue}{rgb}{0.25,0.5,0.5}
\definecolor{codekw}{rgb}{0.85, 0.18, 0.50}
\lstset{
  backgroundcolor=\color{white},
  basicstyle=\fontsize{7.5pt}{7.5pt}\ttfamily\selectfont,
  columns=fullflexible,
  breaklines=true,
  captionpos=b,
  commentstyle=\fontsize{7.5pt}{7.5pt}\color{codeblue},
  keywordstyle=\fontsize{7.5pt}{7.5pt}\color{codekw},
}
\begin{lstlisting}[language=python]
# M: text encoder

for x in loader:  # Load a minibatch x with M samples. The input is assumed to be in a prompt.
    outputs = M(x) # Compute embeddings

    sequence_output, pooled_output = outputs[:2]
    
    sequence_mask_output = sequence_output[:, mask_pos] # Extract masked-token representations

    prediction_mask_scores = lm_head(sequence_mask_output)[:, label_word_list] # Compute logits

    f = get_features_before_last_layer(sequence_mask_output, lm_head) # Features before decoder head 
    prototypes = lm_head.decoder.weight[label_word_list] 
    
    f = normalize(f, dim=1)
    prototypes = normalize(prototypes, dim=1)
    
    sim_mat = f @ prototypes.T #  M-by-K
    
    L_transport = compute_transport_loss(sim_mat, T)
    L_mi = compute_mi_loss(prediction_mask_scores)
    L = L_transport + lambda_mi * L_mi # Loss

    L.backward()  # Back-propagate
    update(M)  # SGD update 

def get_features_before_last_layer(features, lm_head):
    x = lm_head.dense(features)
    x = gelu(x)
    x = lm_head.layer_norm(x)
    return x

\end{lstlisting}
\end{algorithm}

\end{document}